\documentclass[11pt, a4paper, logo, onecolumn, nonumbering]{googledeepmind}

\pdftrailerid{redacted}

\usepackage{marvosym}   
\usepackage{multirow}     
\usepackage{wrapfig}      
\usepackage{placeins}     

\usepackage{subcaption}   
\usepackage{amsmath}
\usepackage{xspace}
\usepackage[most]{tcolorbox}

\newtcblisting{judgeprompt}[2][]{%
  enhanced,
  breakable,
  listing only,
  listing engine=listings,
  colback=black!2,
  colframe=GoogleBlue!65!black,
  colbacktitle=GoogleBlue!9,
  coltitle=black,
  fonttitle=\bfseries,
  boxrule=0.6pt,
  arc=1.5mm,
  left=1.5mm,
  right=1.5mm,
  top=1mm,
  bottom=1mm,
  title={#2},
  listing options={
    basicstyle=\ttfamily\scriptsize,
    breaklines=true,
    breakatwhitespace=false,
    columns=fullflexible,
    keepspaces=true,
    showstringspaces=false,
    upquote=true
  },
  #1
}

\graphicspath{{figures/}}

\definecolor{GoogleBlue}{HTML}{1A73E8}
\usepackage{xurl}
\usepackage[colorlinks=true, allcolors=GoogleBlue]{hyperref}

\definecolor{tideC1}{RGB}{72, 146, 232}
\definecolor{tideC2}{RGB}{142, 120, 204}
\definecolor{tideC3}{RGB}{197, 103, 124}

\newcommand{\appcontentsline}[2]{%
  \noindent\hyperref[#1]{\textbf{\ref*{#1}}\quad #2}%
  \dotfill\hyperref[#1]{\pageref*{#1}}\par}
\newcommand{\appcontentssubline}[2]{%
  \noindent\hspace*{1.8em}\hyperref[#1]{\ref*{#1}\quad #2}%
  \dotfill\hyperref[#1]{\pageref*{#1}}\par}

\def\NickNamecolor{%
  \textcolor{tideC1}{T}\textcolor{tideC1}{I}%
  \textcolor{tideC2}{D}\textcolor{tideC3}{E}}

\title{\emph{\NickNamecolor}: Task-Isolated Diffusion for Unified Video Editing and Generation}

\usepackage{natbib}
\correspondingauthor{jingyuanchen@zju.edu.cn}
\paperurl={https://LittleWork123.github.io/tide}
\reportnumber{}

\fancypagestyle{firststyle}{%
  \fancyhead[L]{\ifthenelse{\boolean{logo}}{\includegraphics[width=32pt]{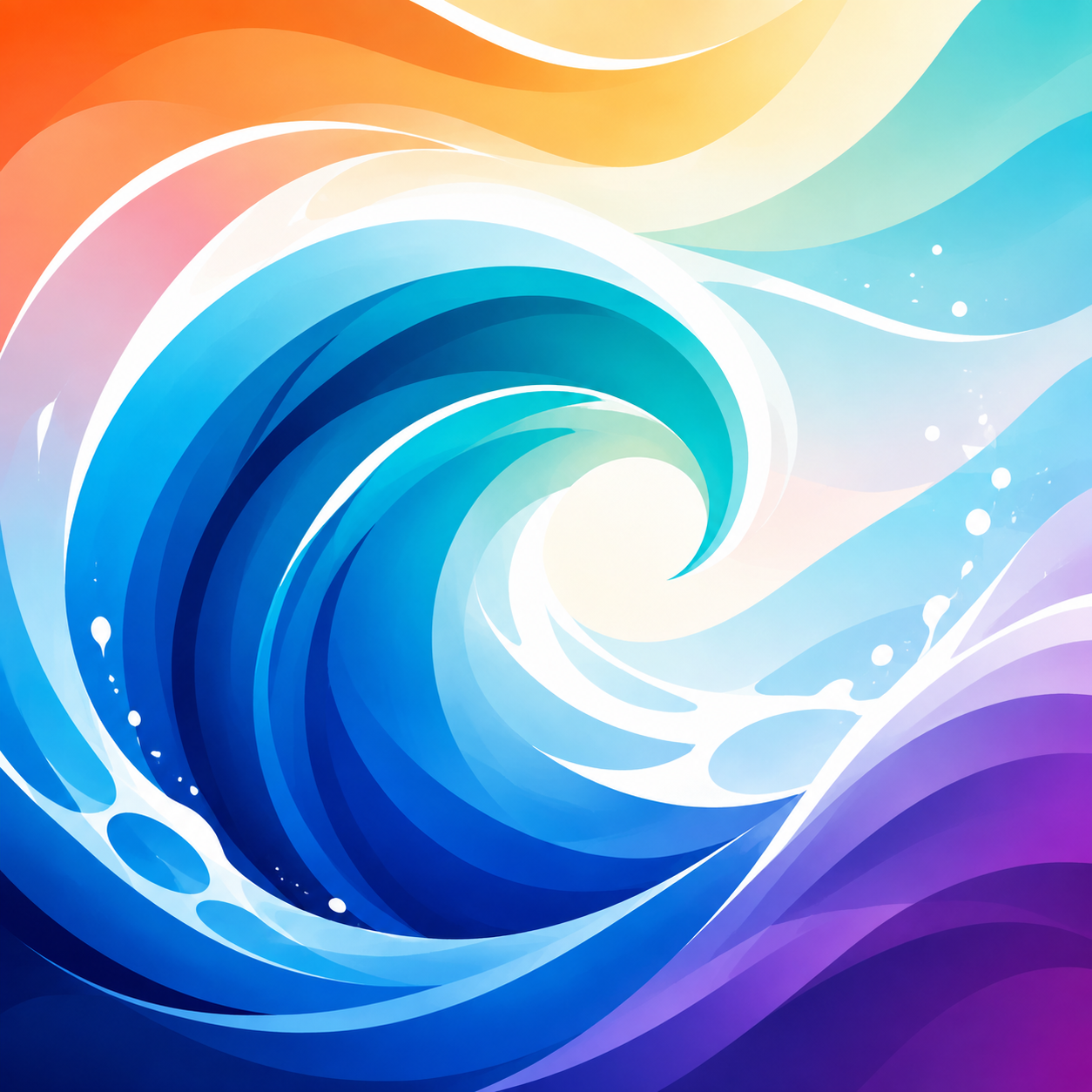}}{}}%
  \fancyhead[R]{%
    \ifdefined\paperurl\if\relax\the\paperurl\relax\else
      \href{\the\paperurl}{\urlheaderfont\itshape\the\paperurl}\\ \fi\fi
    {\footerfont\itshape\monthyeardate\today}}%
  \fancyhead[C]{}%
  \fancyfoot[L]{\footerfont\itshape\textsuperscript{*}~Equal contribution.\quad\textsuperscript{\textdagger}~Corresponding author.}%
  \fancyfoot[R]{}%
  \fancyfoot[C]{}%
}

\author[1,*]{Qi Liu}
\author[2,*]{Gang Yue}
\author[2]{Mingyu Yin}
\author[2]{Lisai Zhang}
\author[2]{Yidi Wu}
\author[2]{Yaole Wang}
\author[2]{Yaohui Wang}
\author[1]{Chang Yao}
\author[1,$\dagger$]{Jingyuan Chen}
\author[2,$\dagger$]{Lin Ma}

\affil[1]{Zhejiang University}
\affil[2]{Bilibili Inc.}

\begin{document}

\begin{abstract}
{\centering
\includegraphics[width=\linewidth]{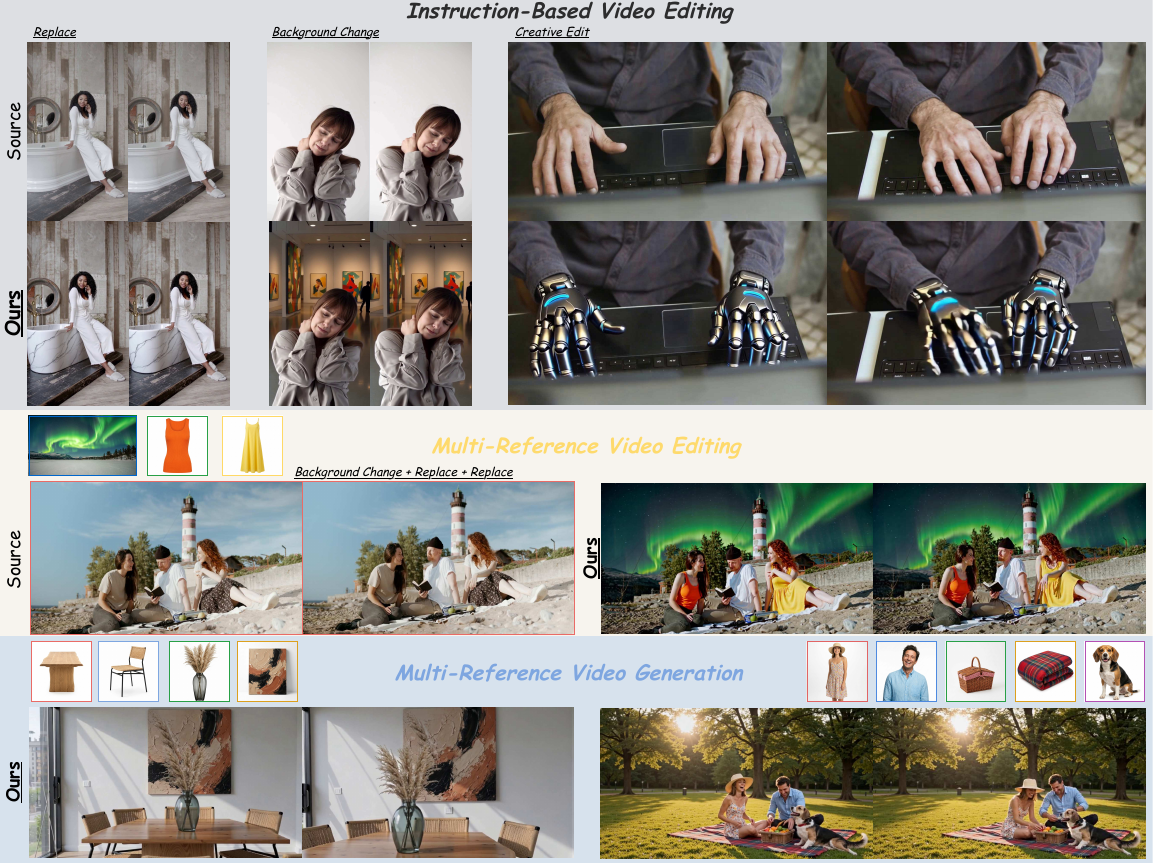}\par}

Recent advances in Diffusion Transformers have driven rapid progress in video generation and editing, yet these capabilities are still handled by separate, task-specific models.
Building a unified framework that supports diverse video tasks remains an open challenge: existing unified attempts either require dedicated auxiliary encoders or lack explicit mechanisms to distinguish heterogeneous conditioning tokens, struggling when the number and type of visual conditions vary across tasks.
We propose TIDE, a unified framework that integrates instruction-based editing, reference-guided editing, and multi-reference generation.
At its core, we introduce per-token task embeddings that assign each input token a task-specific identifier, enabling the model to explicitly disambiguate target, source, and reference tokens.
To simultaneously capture high-level semantic understanding and fine-grained structural fidelity, we design a dual-path conditioning scheme that couples a vision-language model with a VAE latent path for complementary signals.
We further devise a multi-task progressive training strategy that incrementally introduces tasks of increasing complexity, effectively harmonizing diverse objectives and enabling smooth generalization across heterogeneous task distributions.
Extensive experiments demonstrate that TIDE achieves state-of-the-art performance across multiple video editing and generation benchmarks.

\vspace{1em}
\noindent{\includegraphics[height=1em]{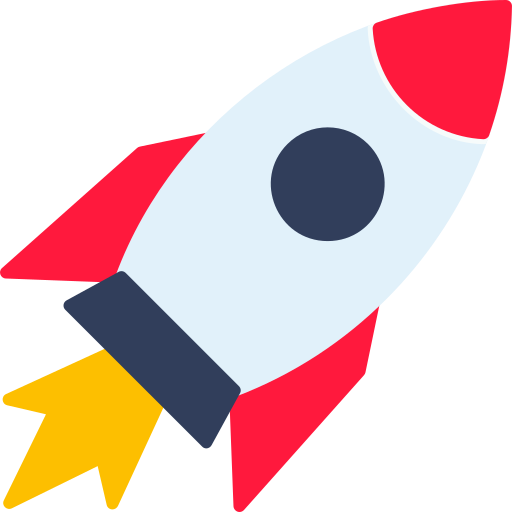}}~\textbf{Project Page:} \url{https://LittleWork123.github.io/tide}\par
\end{abstract}

\newpage
\maketitle

\section{Introduction}
\label{sec:introduction}

The rapid evolution of Diffusion Transformers (DiTs)~\cite{Peebles2022DiT} has propelled video synthesis to a new level, with large-scale models~\cite{wan2025wan,kong2024hunyuanvideo,yang2024cogvideox,hacohen2026ltx} achieving impressive text-to-video and image-to-video generation.
Beyond foundational generation, a proliferation of downstream capabilities has emerged, notably \emph{instruction-based video editing}~\cite{cheng2023consistent,openve,ditto,lucyedit,liao2025context,wu2025insvie,tan2025omni} and \emph{subject-reference video generation}~\cite{liu2025phantom,guo2026dreamidomni}.
However, these capabilities are still handled by separate, task-specific models~\cite{chen2026vino}, and building a unified framework that supports diverse video tasks remains an open challenge.
In the image domain, researchers have begun constructing unified architectures~\cite{xiao2024omnigen,chen2024unireal} that integrate generation, editing, and reference-guided tasks into a cohesive system.
In the video domain, recent works including VACE~\cite{jiang2025vace}, VINO~\cite{chen2026vino}, and DreamID-Omni~\cite{guo2026dreamidomni} represent initial steps toward unified video models, yet they either require dedicated auxiliary encoders or lack explicit mechanisms to prevent inter-task interference.

A closer look reveals that these seemingly disparate tasks can be cast as \emph{a single conditional denoising problem with varying token roles}: the model receives a mixed sequence of target, source, and reference tokens and must learn which to reconstruct, which to preserve, and which to draw identity from.
The key challenge lies in enabling the model to correctly distinguish these roles.
Existing approaches adopt different strategies: VACE~\cite{jiang2025vace} introduces a dedicated context encoder to process conditioning signals separately; VINO~\cite{chen2026vino} uses a structured 3D RoPE layout with special boundary tokens to distinguish heterogeneous visual sources in the attention sequence.
However, these designs either require carefully engineered positional encoding schemes or additional encoder modules, and can struggle when the number of conditioning sources varies (e.g., multi-reference editing with multiple reference images).

Realizing this unified vision, however, requires overcoming three concrete obstacles:
(1)~\textbf{Conditioning disambiguation}: when multiple visual inputs (source video, reference images) are concatenated into a shared attention sequence, the model lacks cues to determine which tokens should be preserved, which provide identity, and which are denoising targets.
VACE~\cite{jiang2025vace} observes that explicitly separating data of different modalities and distributions is essential for model convergence, while VINO~\cite{chen2026vino} reports that without boundary marking, concatenated latents lead to identity swapping and attribute leakage.
(2)~\textbf{Conditioning duality}: reference-guided tasks simultaneously demand high-level semantic understanding (interpreting \emph{what} to generate from text and reference) and fine-grained structural fidelity (preserving \emph{how} it should look), yet no single conditioning pathway can satisfy both~\cite{hu2025hunyuancustom,zhong2025concat}.
(3)~\textbf{Data scarcity and task conflicts}: existing editing datasets~\cite{openve,ditto} provide source-target pairs with instructions but lack reference images, and multi-task training risks inter-task degradation from conflicting learning objectives~\cite{guo2026dreamidomni} without careful curriculum design.

To address these challenges, we introduce \textbf{TIDE}, a unified framework that integrates instruction-based editing, reference-guided editing, and multi-reference generation.
Central to TIDE is a \emph{per-token task embedding}: a learnable embedding table that assigns each input token a task-specific identifier, enabling the model to explicitly distinguish target, source, and reference tokens.
Crucially, the same type of visual input (e.g., a reference image) receives different identifiers depending on the task context, and each reference in multi-reference scenarios is assigned a unique identifier, allowing the model to differentiate multiple visual identities.

Complementing the task isolation, we design a \emph{dual-path conditioning} scheme that combines a semantic path (encoding references jointly with text instructions through Gemma-3-12B-IT~\cite{team2025gemma3} for high-level intent understanding) with a latent path (encoding references through the Video VAE for fine-grained structural and textural detail).
The two paths address the conditioning duality: semantic guidance alone cannot preserve detailed visual appearance~\cite{hu2025hunyuancustom}, while latent conditioning alone lacks the capacity to interpret complex instructions~\cite{zhong2025concat,hacohen2026ltx}.

Finally, to harmonize tasks with varying conditioning strengths, we adopt a \emph{multi-task progressive training} strategy that incrementally introduces tasks of increasing complexity, preventing the model from overfitting to any single task while maintaining strong performance across all tasks.

In summary, our contributions are as follows:
\begin{enumerate}[leftmargin=*,nosep]
    \item We propose TIDE, a unified video editing and generation framework that seamlessly integrates instruction-based editing, reference-guided editing, and multi-reference generation through a shared conditional denoising formulation.
    \item We introduce per-token task embeddings that assign each input token a task-specific identifier, enabling the model to explicitly disambiguate heterogeneous conditioning tokens.
    \item We design a dual-path conditioning scheme that couples VLM-based semantic guidance with VAE-based latent-space detail injection, complemented by a multi-task progressive training strategy that effectively harmonizes tasks with varying conditioning requirements.
    \item Extensive experiments demonstrate that TIDE achieves state-of-the-art performance across multiple video editing and generation benchmarks. We also contribute TIDE-Bench, a new benchmark for evaluating multi-reference video editing, on which TIDE substantially outperforms existing open-source methods.
\end{enumerate}

\section{Related Work}
\label{sec:related_work}

\noindent
\textbf{Video Diffusion Models.}
Early video diffusion models extend image diffusion along the temporal axis using factorized or full 3D U-Nets~\cite{ho2022video,singer2022makeavideo}.
Latent video diffusion subsequently reduces the computational cost by denoising compact spatiotemporal representations~\cite{blattmann2023stable}, while lightweight temporal modules adapt image priors to video generation~\cite{guo2023animatediff}.
More recently, Diffusion Transformers (DiTs)~\cite{Peebles2022DiT} have become the prevailing backbone for scalable video synthesis.
Latte~\cite{ma2024latte} explores latent-space video DiTs, and Sora~\cite{brooks2024sora} demonstrates the scaling potential of treating videos as sequences of spatiotemporal patches.
This design has been adopted by large-scale systems including CogVideoX, Open-Sora, HunyuanVideo, Wan, Movie Gen, and LTX~\cite{yang2024cogvideox,opensora,kong2024hunyuanvideo,wan2025wan,polyak2024movie,hacohen2026ltx}, commonly together with 3D VAEs and flow-matching objectives~\cite{lipman2022flow}.
Although these foundation models provide strong text-to-video and image-to-video priors, they are not by themselves designed to distinguish the heterogeneous source, target, and reference roles required by unified editing and generation.
TIDE builds on LTX-2.3~\cite{hacohen2026ltx} and equips its shared DiT backbone with explicit token-level task isolation.

\noindent
\textbf{Instruction-Based Video Editing.}
Early training-free approaches transfer image-editing priors to video through diffusion inversion, cross-frame attention propagation, or feature correspondence~\cite{song2020denoising,qi2023fatezero,geyer2023tokenflow,kara2024rave}.
These methods avoid task-specific training, but their dependence on inversion quality and internal feature correspondence makes precise local edits and long-term temporal consistency difficult.
Training-based methods instead learn explicit source--instruction--target mappings from paired data.
Large-scale datasets and systems such as OpenVE, Ditto, InsVIE, Lucy-Edit, ICVE, and Omni-Video~\cite{openve,ditto,wu2025insvie,lucyedit,liao2025context,tan2025omni} have substantially improved instruction following, reconstruction fidelity, and edit diversity, while complementary datasets and benchmarks broaden coverage of real-world editing intents~\cite{yu2025veggie,wang2026live,zi2025se,li2025five}.
Recent approaches including Kiwi-Edit~\cite{lin2026kiwiedit}, VINO~\cite{chen2026vino}, and UniVideo~\cite{wei2025univideo} further incorporate multimodal understanding or reference conditions into video editing.
Nevertheless, instruction-only editing typically assumes one source video, whereas reference-guided editing must additionally preserve and bind fine-grained visual evidence from one or more images.

\noindent
\textbf{Reference-Guided and Unified Video Generation and Editing.}
Image-prompt adapters first demonstrate that visual concepts can be injected into text-conditioned diffusion through decoupled cross-attention~\cite{ye2023ip}.
Subject-to-video methods extend this idea to dynamic scenes, progressing from subject-consistent generation in Phantom~\cite{liu2025phantom} to flexible composition and stronger identity preservation in ConceptMaster, BindWeave, LibraGen, Video Alchemist, and Alchemist~\cite{huang2025conceptmaster,li2025bindweave,zhu2026libragen,chen2025videoalchemist,girish2025alchemint}; SkyReels-A2, MVS2V, OmniVCus, MagRef, OpenS2V-Nexus, and HunyuanCustom~\cite{fei2025skyreelsa2,song2026mvs2v,cai2025omnivcus,deng2025magref,yuan2025opens2vnexus,hu2025hunyuancustom} further strengthen multi-reference control through specialized feature extraction and fusion.
Reference-guided editing introduces the additional requirement of selectively transferring reference appearance while preserving the source video's content and motion, addressed through VLM-based multiscale features in MiVE~\cite{wang2026mive}, symmetric conditioning in DreamID-Omni~\cite{guo2026dreamidomni}, and direct latent concatenation~\cite{zhong2025concat}.
In parallel, unified image models such as OmniGen and UniReal~\cite{xiao2024omnigen,chen2024unireal} motivate common formulations for generation and editing, while video models including VACE, VINO, UniVideo, and OmniWeaving~\cite{jiang2025vace,chen2026vino,wei2025univideo,pan2026omniweaving} combine diverse controls or multimodal understanding within shared architectures; Omni-Video~2, Tele-Omni, Many-for-Many, and FullDiT2~\cite{yang2026omnivideo2,liu2026teleomni,yang2025many,he2025fulldit2} further explore generalist conditioning and scalable in-context generation.
However, existing systems typically distinguish heterogeneous inputs through specialized modules, query bottlenecks, positional layouts, or implicit sequence context, making reliable role binding increasingly difficult as the number and type of conditions grow.
TIDE instead assigns lightweight task embeddings directly to target, source, and individual reference tokens, enabling instruction-based editing, reference-guided editing, and single- or multi-reference generation within one shared backbone.

\section{Method}
\label{sec:method}

\begin{figure*}[t]
\centering
\includegraphics[width=0.95\textwidth]{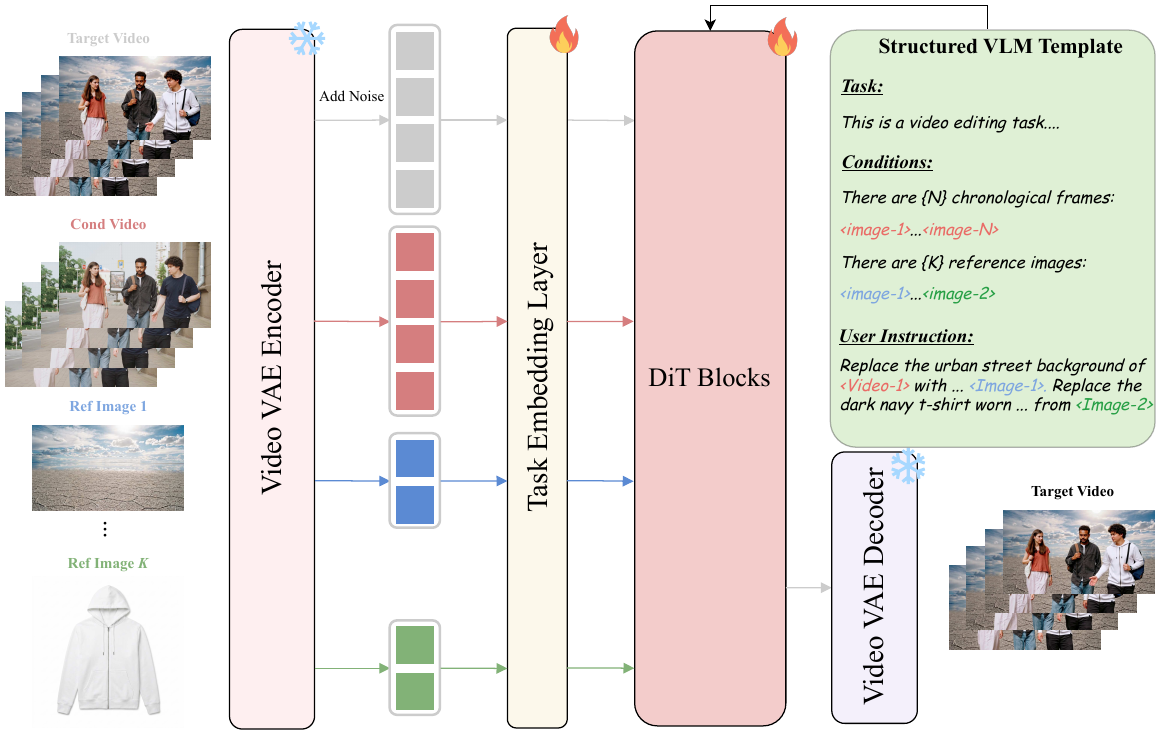}
\caption{\textbf{Overview of TIDE.} Per-token task embeddings isolate heterogeneous conditioning tokens within shared self-attention (\textbf{Left}), while dual-path conditioning combines VLM-based semantic guidance with VAE-encoded latent detail (\textbf{Right}).}
\label{fig:overview}
\end{figure*}

We formalize the problem as a single conditional denoising formulation (\S\ref{sec:formulation}), then describe per-token task embeddings (\S\ref{sec:task_embedding}), dual-path conditioning (\S\ref{sec:dual_path}), and the training procedure (\S\ref{sec:training}).

\subsection{Problem Formulation}
\label{sec:formulation}

We cast reference-based video generation and video editing as a single conditional denoising problem.
Given a text prompt $\mathcal{T}$, optional conditioning visual latents $\mathcal{V} = \{V_1, \dots, V_K\}$ encoded via a 3D VAE, and per-token task identifiers $\boldsymbol{\tau}$, our model learns:
\begin{equation}
    P(Y \mid \mathcal{T}, \mathcal{V}, \boldsymbol{\tau}),
    \label{eq:main}
\end{equation}
where $Y$ denotes the target video.
By selectively providing conditioning signals and assigning appropriate task identifiers, this formulation covers subject-reference video generation, reference-guided video editing, and instruction-based video editing (Table~\ref{tab:task_overview}).

As illustrated in Figure~\ref{fig:overview}, TIDE builds upon LTX-2.3~\cite{hacohen2026ltx} with a 3D Video VAE, a 48-block DiT backbone, and Gemma-3-12B-IT~\cite{team2025gemma3} as the vision-language encoder.
TIDE introduces two mechanisms on top of this backbone: per-token task embeddings (\S\ref{sec:task_embedding}) and dual-path conditioning (\S\ref{sec:dual_path}).

\begin{table}[t]
\centering
\caption{Task unification in TIDE. The same visual input receives \emph{different} task identifiers $\tau$ depending on the task.}
\label{tab:task_overview}
\begin{tabular*}{\textwidth}{@{\extracolsep{\fill}}llll@{}}
\toprule
\textbf{Task} & \textbf{Conditions} & \textbf{Task IDs ($\tau$)} & \textbf{Output} \\
\midrule
Subject-Ref.\ Video Gen. & $\mathcal{T}, \{V_k^{\text{ref}}\}$ & ref: $\tau_1^g, \tau_2^g, \ldots$ & Video \\
Ref-Guided Video Editing & $\mathcal{T}, V_{\text{src}}, \{V_k^{\text{ref}}\}$ & src: $\tau_s^e$,\; ref: $\tau_1^e, \ldots$ & Video \\
Instruction-Based Video Editing & $\mathcal{T}, V_{\text{src}}$ & src: $\tau_s^e$ & Video \\
\bottomrule
\end{tabular*}
\end{table}

\subsection{Per-Token Task Embedding}
\label{sec:task_embedding}

Unifying multiple tasks introduces a fundamental tension: subject-reference generation requires the model to faithfully preserve reference identity, while editing demands selective modification of source content.
Without an explicit signal to distinguish these conditioning roles, the model conflates preservation with modification, leading to inter-task interference.
Existing methods address this via dedicated auxiliary encoders~\cite{jiang2025vace} or carefully designed positional encoding with boundary tokens~\cite{chen2026vino}, but these approaches can be difficult to extend when the number of conditioning sources varies across tasks.

\begin{figure}[t]
\centering
\includegraphics[width=\textwidth]{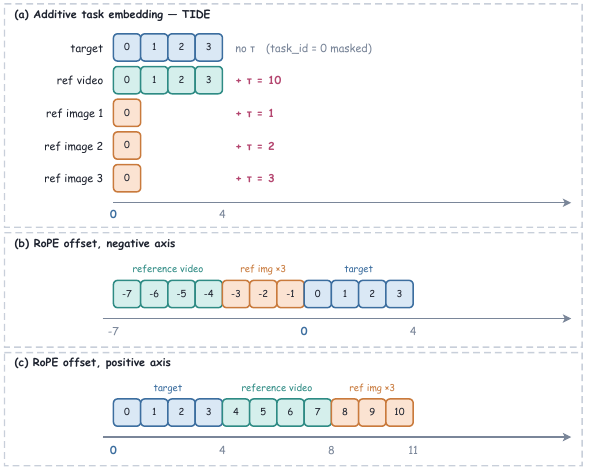}
\caption{Additive task embedding (a) versus two RoPE offset schemes: RoPE-Neg (b) and RoPE-Pos (c). TIDE adds a learned $\tau$ per conditioning source without changing token positions. RoPE-Neg places conditioning tokens on a negative position axis ahead of the target, whereas RoPE-Pos keeps the target starting at zero and appends conditioning tokens on the positive position axis.}
\label{fig:ropevtask}
\end{figure}

We introduce a learnable task embedding table $\mathbf{E} \in \mathbb{R}^{N \times D}$, where $N$ is the number of task slots and $D$ the hidden dimension.
For each patchified latent token $\mathbf{h}_i$ with assigned task identifier $\tau_i$, the task-conditioned representation is:
\begin{equation}
    \tilde{\mathbf{h}}_i = \mathbf{h}_i + \mathbf{E}[\tau_i] \cdot \mathbb{1}[\tau_i \neq 0],
    \label{eq:task_embed}
\end{equation}
where $\mathbf{E}[\tau_i]$ indexes the $\tau_i$-th row.
This design ensures that target tokens ($\tau{=}0$) receive zero task-embedding perturbation, leaving their positional and task-embedding representation unchanged, while each conditioning token receives a learned embedding that signals its role to the attention mechanism.
Figure~\ref{fig:ropevtask} contrasts this design with two alternatives that encode token roles positionally: RoPE-Neg places conditioning tokens on a negative position axis ahead of the target, whereas RoPE-Pos keeps the target starting at zero and appends conditioning tokens on the positive position axis.
The positional scheme separates sources but ties their identity to position, so the role a token plays depends on where it sits in the sequence; the additive scheme carries the role in the token itself, leaving the target's positional layout untouched.
We compare the two empirically in \S\ref{sec:ablation}.

Identifiers are assigned at the \emph{token level} and partitioned into contiguous ranges for different conditioning roles.
For subject-reference generation, each reference image receives a unique identifier ($\tau^g_1, \tau^g_2, \ldots$); for editing tasks, the source video and reference images receive identifiers from a separate range ($\tau^e_s$ for source, $\tau^e_1, \tau^e_2, \ldots$ for references).
Crucially, the same type of visual input receives \emph{different} identifiers depending on the task context: a reference image used in subject-to-video generation and the same image used in reference-guided editing are assigned different identifiers, enabling the model to learn task-appropriate behavior for each conditioning role.
For multi-reference tasks, each reference occupies a unique identifier, allowing the model to differentiate multiple visual identities.
Extending to new tasks requires only allocating new identifiers and providing corresponding training data.

\subsection{Dual-Path Conditioning}
\label{sec:dual_path}

Reference-guided tasks simultaneously demand high-level semantic understanding (interpreting \emph{what} to generate from text and reference) and fine-grained structural fidelity (preserving \emph{how} it should look).
No single conditioning pathway satisfies both~\cite{hu2025hunyuancustom,zhong2025concat}: VLM-based encoding excels at capturing semantic intent but compresses fine visual details, while VAE-based latent encoding retains structural information but cannot interpret complex instructions.
We therefore design a dual-path scheme that injects complementary signals through two channels.

\noindent\textbf{Semantic Path.}
The VLM path jointly encodes visual conditions with the text instruction through Gemma-3-12B-IT, producing contextualized embeddings that capture the semantic relationship between visual inputs and the generation or editing intent.
We design task-specific structured prompt templates (illustrated in Figure~\ref{fig:overview}).
For subject-reference generation, reference images are interleaved with descriptive anchors within the VLM input.
For editing tasks, $N$ frames uniformly sampled from the source video ($N{=}5$ by default) replace the reference slots to provide temporal context, and for reference-guided editing, subject reference images are additionally appended after the source frames.
The VLM outputs are projected through the Embeddings Connector and serve as keys and values for cross-attention in each DiT block, providing high-level semantic guidance that is difficult to derive from latent-space conditioning alone.

\noindent\textbf{Latent Path.}
The latent path encodes all visual inputs through the 3D Video VAE and concatenates them into a single token sequence within the self-attention.
For reference-guided editing, the complete sequence takes the form:
\begin{equation}
    \mathbf{X} = [\underbrace{\mathbf{H}_{\text{target}}}_{\tau=0};\; \underbrace{\mathbf{H}_{\text{src}}}_{\tau_s^e};\; \underbrace{\mathbf{H}_{\text{ref}_1}}_{\tau_1^e};\; \underbrace{\mathbf{H}_{\text{ref}_2}}_{\tau_2^e};\; \ldots\;],
    \label{eq:concat}
\end{equation}
where $[;\;]$ denotes sequence concatenation.
For subject-reference generation, the source term is absent and reference tokens receive generation-specific identifiers ($\tau_k^g$ instead of $\tau_k^e$).
Each conditioning source receives its dedicated task identifier from Eq.~\eqref{eq:task_embed}, enabling the model to distinguish tokens from different sources.
Unlike prior approaches that rely on shared concatenated contexts or dedicated auxiliary encoders~\cite{hacohen2026ltx,guo2026dreamidomni,jiang2025vace}, TIDE directly assigns an explicit role identifier to every visual token.
A binary conditioning mask $\mathbf{m} \in \{0, 1\}^L$ keeps conditioning tokens noise-free ($\sigma_i{=}0$) while target tokens receive diffusion noise ($\sigma_i{=}\sigma$), and the training loss is computed exclusively over target tokens.

Together, the two paths form a complementary conditioning system: the semantic path provides high-level understanding of editing intent via cross-attention, guiding \emph{what} to generate, while the latent path preserves fine-grained structural detail via self-attention with task isolation, ensuring \emph{how} it should look while preventing interference between conditioning sources.
Ablation studies in \S\ref{sec:ablation} confirm the importance of the VLM semantic path and of incorporating source-video semantics into this path.

\subsection{Training and Inference}
\label{sec:training}

\noindent\textbf{Multi-Task Progressive Training.}
Naively mixing all tasks from the start risks inter-task degradation, as editing objectives can conflict with generation objectives during early optimization~\cite{guo2026dreamidomni}.
We adopt a three-stage progressive strategy that incrementally introduces tasks of increasing complexity.
Stage~1 trains exclusively on instruction-based video editing to establish basic video-editing capability.
Stage~2 introduces the full multi-task mixture, including reference-guided editing, subject-to-video generation, and multi-reference tasks, to warm up multi-task representations.
Stage~3 continues multi-task training with refined sampling ratios and a reduced learning rate for long-horizon convergence.
This progressive curriculum allows the model to first consolidate basic editing competence before adapting to the more challenging multi-reference scenarios.
Detailed step counts, learning rates, and data ratios are reported in \S\ref{sec:setup} and the supplementary material.

\noindent\textbf{Objective.}
We adopt flow matching~\cite{lipman2022flow} with shifted logit-normal timestep sampling.
Given a clean sample $\mathbf{x}_0$ and noise $\boldsymbol{\epsilon} \sim \mathcal{N}(0, \mathbf{I})$, the noisy sample at timestep $\sigma$ is $\mathbf{x}_\sigma = (1 - \sigma)\mathbf{x}_0 + \sigma\boldsymbol{\epsilon}$.
The model predicts the velocity field $\mathbf{v} = \boldsymbol{\epsilon} - \mathbf{x}_0$, and the training loss is:
\begin{equation}
    \mathcal{L} = \mathbb{E}_{\sigma, \boldsymbol{\epsilon}} \left[ \frac{1}{|\mathcal{S}|} \sum_{i \in \mathcal{S}} \left\| \mathbf{v}_\theta(\mathbf{x}_\sigma, \mathbf{c}, \boldsymbol{\tau}, \sigma)_i - (\boldsymbol{\epsilon} - \mathbf{x}_0)_i \right\|^2 \right],
    \label{eq:loss}
\end{equation}
where $\mathcal{S} = \{i : \mathbf{m}_i = 0\}$ denotes target token indices, ensuring that the model is trained only to reconstruct the target video while leaving conditioning tokens unchanged.

\noindent\textbf{Inference.}
At inference time, we employ classifier-free guidance (CFG) combined with spatiotemporal guidance (STG)~\cite{hong2024stg}:
\begin{equation}
    \hat{\mathbf{v}} = \mathbf{v}_{\text{uncond}} + s_{\text{cfg}} (\mathbf{v}_{\text{cond}} - \mathbf{v}_{\text{uncond}}) + s_{\text{stg}} (\mathbf{v}_{\text{cond}} - \mathbf{v}_{\text{stg}}),
    \label{eq:full_guidance}
\end{equation}
where $s_{\text{cfg}}{=}4.0$ and $s_{\text{stg}}{=}1.0$.
We use a 50-step Euler schedule, generating videos at $1280{\times}704$ resolution.

\section{Experiments}
\label{sec:experiments}

\begin{figure*}[t]
\centering
\includegraphics[width=1.0\textwidth]{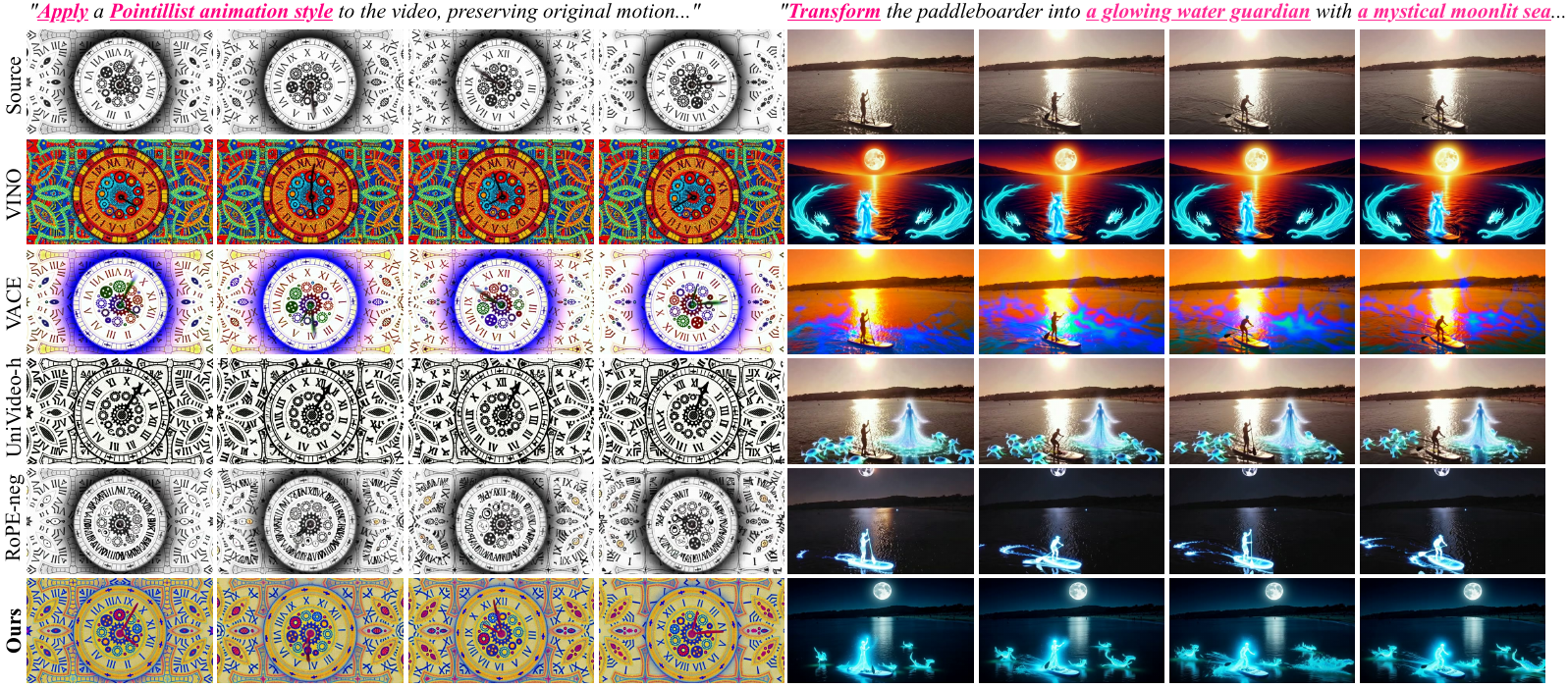}
\caption{\textbf{Qualitative comparison} on OpenVE-Bench.}
\label{fig:qualitative_edit}
\end{figure*}

\begin{figure*}[t]
\centering
\includegraphics[width=1.0\textwidth]{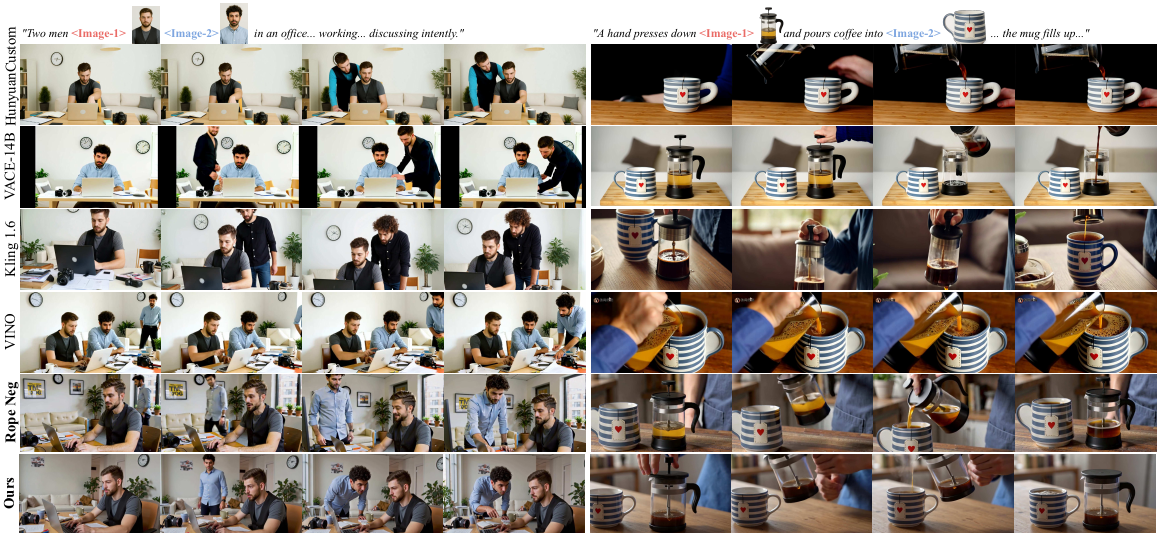}
\caption{\textbf{Qualitative comparison} on multi-reference subject-to-video generation.}
\label{fig:qualitative_gen}
\end{figure*}

\subsection{Experimental Setup}
\label{sec:setup}

\noindent\textbf{Implementation Details.}
We build TIDE on top of LTX-2.3~\cite{hacohen2026ltx}, using its native Gemma-3-12B-IT~\cite{team2025gemma3} as the VLM encoder (frozen throughout training) and its 14B-parameter DiT backbone, which is fully fine-tuned together with randomly initialized task embedding tables.
Training proceeds through three progressive stages (Stage~1: ${\sim}$3K steps, Stage~2: ${\sim}$7K steps, Stage~3: ${\sim}$10K steps) on H20 GPUs with FSDP; detailed hyperparameters, data ratios, and per-category data counts are provided in the supplementary material.
At inference, we use CFG with $s_{\text{cfg}}{=}4.0$.
For generation tasks, we produce 145-frame videos at 24~fps and $1280 \times 704$ resolution; for editing tasks, the frame count and resolution follow the source video from each benchmark.

\noindent\textbf{Benchmarks and Metrics.}
We evaluate on six benchmark tasks spanning video editing and generation.
\textbf{OpenVE-Bench}~\cite{openve} covers eight instruction-based video editing categories.
\textbf{TIDE-Bench} (ours) evaluates multi-reference video editing with single and compound operations through Edit Completeness, Reference Faithfulness, Visual \& Temporal Quality, Scene Preservation, and Overall Quality.
Both editing benchmarks are scored by Gemini-3.1-Pro~\cite{gemini31pro}.
\textbf{OpenS2V}~\cite{yuan2025opens2vnexus} evaluates subject-to-video generation using Aesthetics, Motion Smoothness, Motion Amplitude, GME, Nexus, Naturalness, and their weighted Total.
\textbf{RefVIE-Bench}~\cite{lin2026kiwiedit} covers subject- and background-reference-guided editing.
Finally, we adopt the \textbf{TIV2V} reference-guided editing and \textbf{Compositional MI2V} multi-reference generation tasks of IntelligentVBench~\cite{pan2026omniweaving}.
For RefVIE-Bench and IntelligentVBench, we follow the official protocols and use Gemini-2.5-Pro as the judge on a 1--5 scale.

\noindent\textbf{Baselines.}
For instruction-based editing (OpenVE-Bench), we compare with open-source methods InsVIE~\cite{wu2025insvie}, DITTO~\cite{ditto}, Lucy-Edit~\cite{lucyedit}, ICVE~\cite{liao2025context}, Omni-Video~\cite{tan2025omni}, UniVideo~\cite{wei2025univideo} (both its query- and hidden-state conditioning variants), Kiwi-Edit~\cite{lin2026kiwiedit}, VINO~\cite{chen2026vino}, and closed-source Kling-O1~\cite{team2025kling} and SkyReels-Omni~\cite{chen2026skyreelsv4}.
For multi-reference editing (TIDE-Bench), we compare with OmniWeaving~\cite{pan2026omniweaving}, VINO~\cite{chen2026vino}, Kiwi-Edit~\cite{lin2026kiwiedit}, and both UniVideo~\cite{wei2025univideo} variants, along with closed-source Kling-O1~\cite{team2025kling} and SkyReels-Omni~\cite{chen2026skyreelsv4}.
The released TIV2V inference interface of OmniWeaving used in our evaluation accepts one reference image, so its result is reported on the single-reference subset of TIDE-Bench; all other methods are evaluated on the full benchmark.
For subject-to-video generation (OpenS2V), we compare with MagRef~\cite{deng2025magref}, Phantom~\cite{liu2025phantom}, SkyReels-A2~\cite{fei2025skyreelsa2}, VACE~\cite{jiang2025vace}, HunyuanCustom~\cite{hu2025hunyuancustom}, VINO~\cite{chen2026vino}, and closed-source Kling~1.6~\cite{team2025kling}, Pika~2.1~\cite{pika2025}, and Vidu~2.0~\cite{vidu2025}.
For RefVIE-Bench and IntelligentVBench, we follow their published comparisons against Kiwi-Edit, VINO, UniVideo, OmniWeaving, VACE, and specialized subject-driven generation models as applicable to each task.

\subsection{Quantitative Results}
\label{sec:exp_results}

\noindent\textbf{Results on OpenVE-Bench.}
For instruction-based video editing, we evaluate TIDE on OpenVE-Bench, with results reported in Table~\ref{tab:openve}.
TIDE achieves the highest average score among open-source methods, improving over the second-best VINO from 2.60 to 2.91, and ranks first on five of the eight editing categories.
This consistent performance across both global and local edits demonstrates that the unified model can accurately interpret diverse editing intents while preserving strong editing quality.
Moreover, its clear advantage over RoPE-Neg and RoPE-Pos (2.37 and 2.24) verifies that learned task embeddings provide a more effective interface for heterogeneous visual conditions.

\begin{table}[t]
\centering
\caption{Comparison with SOTA methods on OpenVE-Bench. The \textbf{best} and \underline{second-best} results are marked among open-source methods.}
\label{tab:openve}
\scalebox{0.85}{
\setlength{\tabcolsep}{3.5pt}
\begin{tabular*}{1.18\textwidth}{@{\extracolsep{\fill}}lcccccccc c@{}}
\toprule
\textbf{Method} & \textbf{Style} & \textbf{BG} & \textbf{Chg} & \textbf{Rm} & \textbf{Add} & \textbf{Sub} & \textbf{Cre} & \textbf{Cam} & \textbf{Avg.} \\
\midrule
\multicolumn{10}{l}{\textit{Closed-source}} \\
Kling-O1~\cite{team2025kling} & 4.32 & 2.44 & 4.01 & 3.03 & 2.89 & 3.12 & 3.44 & 3.75 & 3.36 \\
SkyReels-Omni~\cite{chen2026skyreelsv4} & 4.41 & 2.23 & 4.19 & 3.35 & 2.36 & 3.62 & 3.44 & 1.23 & 3.14 \\
\midrule
\multicolumn{10}{l}{\textit{Open-source}} \\
InsVIE~\cite{wu2025insvie} & 1.40 & 1.00 & 1.29 & 1.00 & 1.04 & 1.17 & 1.50 & 1.04 & 1.16 \\
DITTO~\cite{ditto} & 3.72 & 1.09 & 1.87 & 1.00 & 1.39 & 1.09 & 2.35 & 1.05 & 1.68 \\
Lucy-Edit~\cite{lucyedit} & 2.33 & 1.33 & 2.80 & 1.14 & 2.01 & 1.06 & 2.46 & 1.00 & 1.78 \\
ICVE~\cite{liao2025context} & 2.40 & 1.18 & 2.23 & 1.85 & 1.75 & 2.24 & 2.09 & 1.01 & 1.84 \\
Omni-Video~\cite{tan2025omni} & 3.07 & 1.11 & 2.37 & 1.39 & 1.87 & 2.33 & \underline{2.92} & \underline{1.77} & 2.05 \\
UniVideo (query)~\cite{wei2025univideo} & 3.73 & 1.80 & 2.78 & 1.73 & \textbf{2.26} & 2.71 & 2.52 & 1.06 & 2.35 \\
UniVideo (hidden)~\cite{wei2025univideo} & 3.73 & 2.05 & 3.04 & 2.40 & \underline{2.20} & 2.53 & 2.90 & 1.04 & 2.51 \\
Kiwi-Edit~\cite{lin2026kiwiedit} & 3.61 & \underline{2.08} & \underline{3.07} & \textbf{2.84} & 2.16 & 2.31 & 2.72 & 1.07 & 2.52 \\
VINO~\cite{chen2026vino} & \underline{4.31} & 1.50 & 2.94 & 2.38 & 2.07 & \underline{2.78} & 2.63 & \textbf{2.22} & \underline{2.60} \\
\midrule
TIDE w/ RoPE-Neg & 3.16 & 1.76 & 2.92 & 2.42 & 1.96 & \underline{2.78} & 2.62 & 1.26 & 2.37 \\
TIDE w/ RoPE-Pos & 3.03 & 1.74 & 2.73 & 2.54 & 1.99 & 2.07 & 2.47 & 1.12 & 2.24 \\
\rowcolor{gray!10}
\textbf{TIDE (Ours)} & \textbf{4.32} & \textbf{2.62} & \textbf{3.54} & \underline{2.68} & 2.18 & \textbf{3.56} & \textbf{2.93} & 1.15 & \textbf{2.91} \\
\bottomrule
\end{tabular*}
}
\end{table}

\noindent\textbf{Results on TIDE-Bench.}
We further evaluate multi-reference video editing on TIDE-Bench, as shown in Table~\ref{tab:tidebench}.
TIDE achieves the highest average score among all open-source methods, outperforming the second-place UniVideo (hidden) by a substantial margin of 0.66, and obtains the best score on every evaluation dimension.
These results highlight TIDE's ability to execute compound editing instructions, faithfully bind multiple references to their intended targets, and preserve the source scene without sacrificing temporal or visual quality.
The improvement over RoPE-Neg and RoPE-Pos (3.41 versus 2.97 and 2.66) further demonstrates the importance of explicitly identifying the semantic role of each visual input.

\begin{table}[t]
\centering
\caption{Comparison with SOTA methods on TIDE-Bench. All open-source baselines are evaluated by us using the same judge and scoring protocol, and the \textbf{best} and \underline{second-best} results are marked among open-source methods. The released OmniWeaving TIV2V inference interface used in our evaluation accepts one reference image, so its score is computed on the single-reference subset; all other methods are evaluated on the full benchmark.}
\label{tab:tidebench}
\scalebox{0.82}{
\begin{tabular*}{1.22\textwidth}{@{\extracolsep{\fill}}lcccccc@{}}
\toprule
\textbf{Method} & \textbf{Edit$\uparrow$} & \textbf{Ref$\uparrow$} & \textbf{V\&T$\uparrow$} & \textbf{Pres$\uparrow$} & \textbf{Qual$\uparrow$} & \textbf{Avg.$\uparrow$} \\
\midrule
\multicolumn{7}{l}{\textit{Closed-source}} \\
Kling-O1~\cite{team2025kling} & 4.09 & 3.78 & 3.43 & 3.47 & 3.40 & 3.63 \\
SkyReels-Omni~\cite{chen2026skyreelsv4} & 4.29 & 4.01 & 3.58 & 3.63 & 3.53 & 3.81 \\
\midrule
\multicolumn{7}{l}{\textit{Open-source}} \\
OmniWeaving~\cite{pan2026omniweaving} & 2.68 & 2.59 & 2.07 & 2.12 & 1.97 & 2.29 \\
VINO~\cite{chen2026vino} & 3.05 & 2.73 & 2.27 & 2.14 & 2.07 & 2.45 \\
Kiwi-Edit~\cite{lin2026kiwiedit} & 3.22 & 2.43 & 2.43 & 2.44 & 2.31 & 2.56 \\
UniVideo (query)~\cite{wei2025univideo} & 2.83 & 3.18 & 2.28 & 2.33 & 2.21 & 2.57 \\
UniVideo (hidden)~\cite{wei2025univideo} & 3.18 & \underline{3.37} & 2.43 & 2.48 & 2.32 & 2.75 \\
\midrule
TIDE w/ RoPE-Neg & \underline{3.59} & 3.00 & \underline{2.79} & \underline{2.80} & \underline{2.68} & \underline{2.97} \\
TIDE w/ RoPE-Pos & 3.33 & 2.71 & 2.45 & 2.47 & 2.36 & 2.66 \\
\rowcolor{gray!10}
\textbf{TIDE (Ours)} & \textbf{4.07} & \textbf{3.44} & \textbf{3.20} & \textbf{3.21} & \textbf{3.15} & \textbf{3.41} \\
\bottomrule
\end{tabular*}
}
\end{table}

\begin{figure*}[t]
\centering
\includegraphics[width=\textwidth]{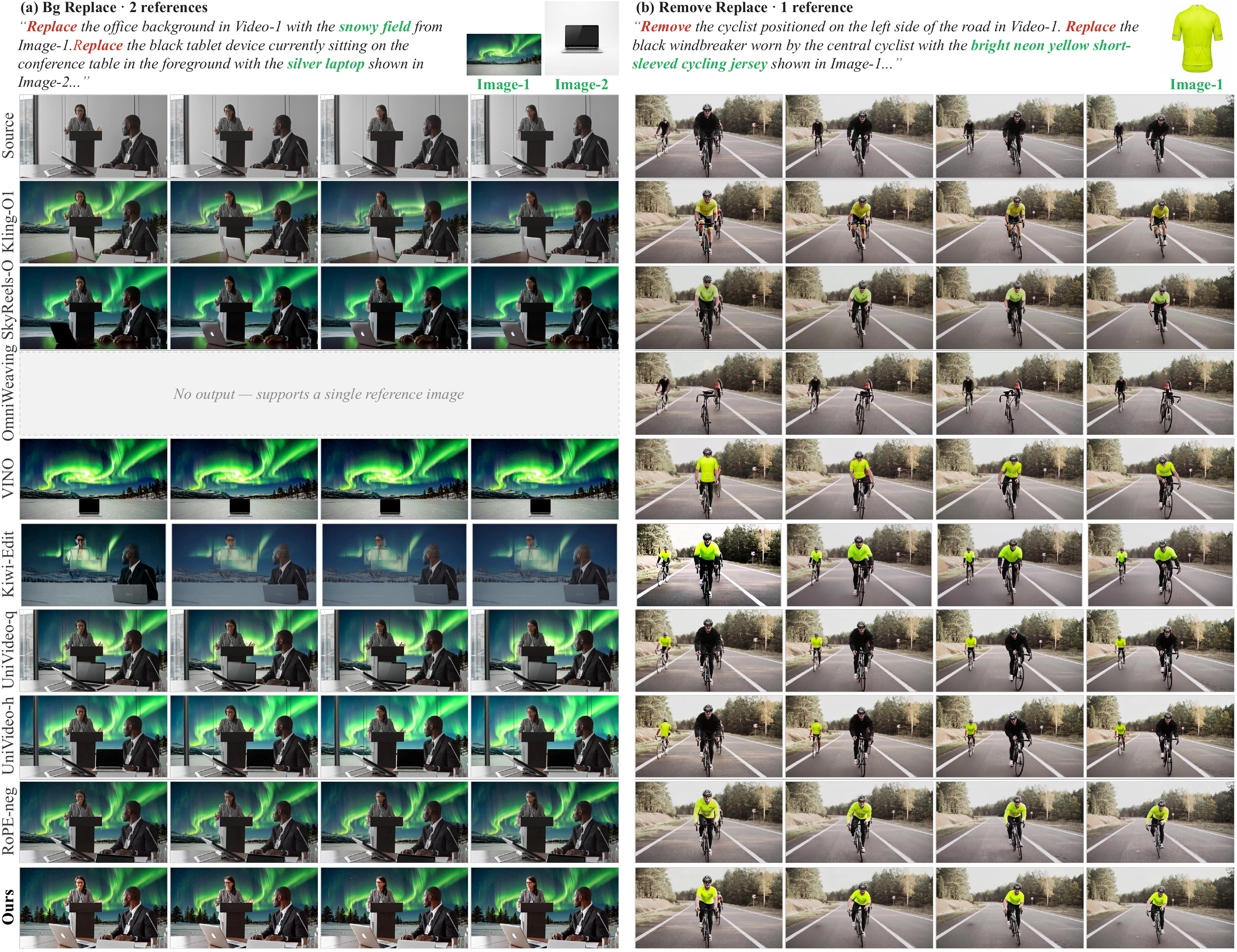}
\caption{\textbf{Qualitative comparison} on TIDE-Bench.}
\label{fig:tidebench_qualitative}
\end{figure*}

\noindent\textbf{Results on OpenS2V.}
For subject-to-video generation, we report results on OpenS2V in Table~\ref{tab:opens2v}.
TIDE achieves the highest overall Total score of 62.62, surpassing the closed-source Kling~1.6 by 2.36 points and the strongest open-source baseline VINO by 3.31 points.
It also obtains the best open-source Aesthetics and Naturalness scores together with a competitive Nexus score, demonstrating that our unified training strategy preserves reference identity and prompt alignment while maintaining favorable visual quality and physical plausibility.
RoPE-Neg and RoPE-Pos obtain Total scores of 59.70 and 59.20, respectively; their lower aggregate performance further supports the effectiveness of learned task embeddings for subject-driven generation.

\begin{table}[t]
\centering
\caption{Comparison with SOTA methods on OpenS2V. The \textbf{best} and \underline{second-best} results are marked among open-source methods.}
\label{tab:opens2v}
\scalebox{0.82}{
\setlength{\tabcolsep}{3pt}
\begin{tabular*}{1.22\textwidth}{@{\extracolsep{\fill}}lccccccc@{}}
\toprule
\textbf{Method} & \textbf{Total$\uparrow$} & \textbf{Aes$\uparrow$} & \textbf{MSmooth$\uparrow$} & \textbf{MAmp$\uparrow$} & \textbf{GME$\uparrow$} & \textbf{Nexus$\uparrow$} & \textbf{Natural$\uparrow$} \\
\midrule
\multicolumn{8}{l}{\textit{Closed-source}} \\
Kling 1.6~\cite{team2025kling} & 60.26 & 44.59 & 86.93 & 41.60 & 66.20 & 45.89 & 74.59 \\
Pika 2.1~\cite{pika2025} & 57.25 & 46.88 & 87.06 & 24.71 & 69.19 & 45.40 & 63.32 \\
Vidu 2.0~\cite{vidu2025} & 56.16 & 41.48 & 90.45 & 13.52 & 67.57 & 43.37 & 65.88 \\
\midrule
\multicolumn{8}{l}{\textit{Open-source}} \\
MAGREF~\cite{deng2025magref} & 57.94 & 45.02 & 93.17 & 21.81 & 70.47 & 43.04 & 66.90 \\
Phantom-1.3B~\cite{liu2025phantom} & 56.47 & 46.67 & 93.30 & 14.29 & 69.43 & 42.48 & 62.50 \\
Phantom-14B~\cite{liu2025phantom} & 58.10 & 46.39 & 96.31 & \textbf{33.42} & 70.65 & 37.43 & 69.35 \\
SkyReels-A2~\cite{fei2025skyreelsa2} & 53.83 & 39.41 & 87.93 & \underline{25.60} & 64.54 & 43.75 & 60.32 \\
VACE-1.3B~\cite{jiang2025vace} & 57.21 & 48.24 & \textbf{97.20} & 18.83 & \underline{71.26} & 37.91 & 65.46 \\
VACE-14B~\cite{jiang2025vace} & 58.16 & 47.21 & 94.97 & 15.02 & 67.27 & 44.08 & 67.04 \\
VACE-P1.3B~\cite{jiang2025vace} & 57.08 & 47.34 & \underline{96.80} & 12.03 & \textbf{71.38} & 40.19 & 64.31 \\
HunyuanCustom~\cite{hu2025hunyuancustom} & 57.69 & \underline{48.26} & 96.37 & 6.93 & 66.11 & 43.56 & 65.24 \\
VINO~\cite{chen2026vino} & 59.31 & 45.92 & 94.73 & 12.30 & 69.69 & 42.67 & \underline{71.99} \\
\midrule
TIDE w/ RoPE-Neg & \underline{59.70} & 45.38 & 93.80 & 17.38 & 66.48 & 46.77 & 71.63 \\
TIDE w/ RoPE-Pos & 59.20 & 45.48 & 88.95 & 12.23 & 64.40 & \underline{47.24} & 64.77 \\
\rowcolor{gray!10}
\textbf{TIDE (Ours)} & \textbf{62.62} & \textbf{49.53} & 95.84 & 22.56 & 70.09 & \textbf{49.40} & \textbf{73.66} \\
\bottomrule
\end{tabular*}
}
\end{table}

\noindent\textbf{Results on RefVIE-Bench.}
We evaluate instruction-and-reference-guided video editing with subject or background references on RefVIE-Bench.
As presented in Table~\ref{tab:refvie}, TIDE achieves the highest overall score among all open-source methods, improving over the second-best VINO from 3.53 to 3.89, a relative gain of 10.2\%.
TIDE also ranks first across all six dimensions, demonstrating robust identity preservation and temporal consistency for subject replacement as well as accurate reference restoration and visual harmony for background replacement.
Its substantial advantage over RoPE-Neg and RoPE-Pos (3.89 versus 3.28 and 2.79) confirms that explicit task embeddings enable more precise control over fine-grained visual conditions.

\begin{table}[h]
\centering
\caption{Quantitative comparison on RefVIE-Bench (Gemini-2.5-Pro). Overall is the sample-weighted average over the 120 samples (80 subject, 40 background). The \textbf{best} and \underline{second-best} scores are compared among open-source models only.}
\label{tab:refvie}
\resizebox{\textwidth}{!}{%
\begin{tabular}{l|ccc|ccc|c}
\toprule
\multirow{2}{*}{\textbf{Model}} & \multicolumn{3}{c|}{\textbf{Subject Reference}} & \multicolumn{3}{c|}{\textbf{Background Reference}} & \multirow{2}{*}{\textbf{Overall$\uparrow$}} \\
 & \textbf{Identity$\uparrow$} & \textbf{Temporal$\uparrow$} & \textbf{Physical$\uparrow$} & \textbf{Ref.\ Sim$\uparrow$} & \textbf{Matting$\uparrow$} & \textbf{Quality$\uparrow$} & \\
\midrule
\multicolumn{8}{l}{\textit{Closed-Source Models}} \\
Runway Aleph & 3.79 & 3.65 & 3.58 & 3.33 & 2.81 & 2.58 & 3.29 \\
Kling-O1 & 4.75 & 4.66 & 4.60 & 3.95 & 3.21 & 2.75 & 3.99 \\
\midrule
\multicolumn{8}{l}{\textit{Open-Source Models}} \\
Kiwi-Edit (All data) & 3.51 & 2.96 & 2.91 & 3.40 & 2.58 & 2.40 & 2.96 \\
Kiwi-Edit (Ref.\ data only) & 3.98 & 3.40 & 3.34 & \underline{3.72} & \underline{2.90} & 2.51 & 3.31 \\
VINO & 4.18 & \underline{4.03} & \underline{3.74} & 2.93 & 2.60 & 2.40 & \underline{3.53} \\
UniVideo & \underline{4.19} & 3.80 & 3.61 & 2.90 & 2.22 & 2.12 & 3.38 \\
OmniWeaving & 3.29 & 2.96 & 2.82 & 3.45 & 2.55 & 2.35 & 2.94 \\
\midrule
TIDE w/ RoPE-Neg & 3.69 & 3.38 & 3.23 & 3.50 & 2.80 & \underline{2.60} & 3.28 \\
TIDE w/ RoPE-Pos & 3.15 & 2.89 & 2.79 & 2.88 & 2.33 & 2.23 & 2.79 \\
\textbf{TIDE (Ours)} & \textbf{4.38} & \textbf{4.15} & \textbf{4.01} & \textbf{3.90} & \textbf{3.15} & \textbf{2.90} & \textbf{3.89} \\
\bottomrule
\end{tabular}
}
\end{table}

\begin{figure*}[t]
\centering
\includegraphics[width=\textwidth]{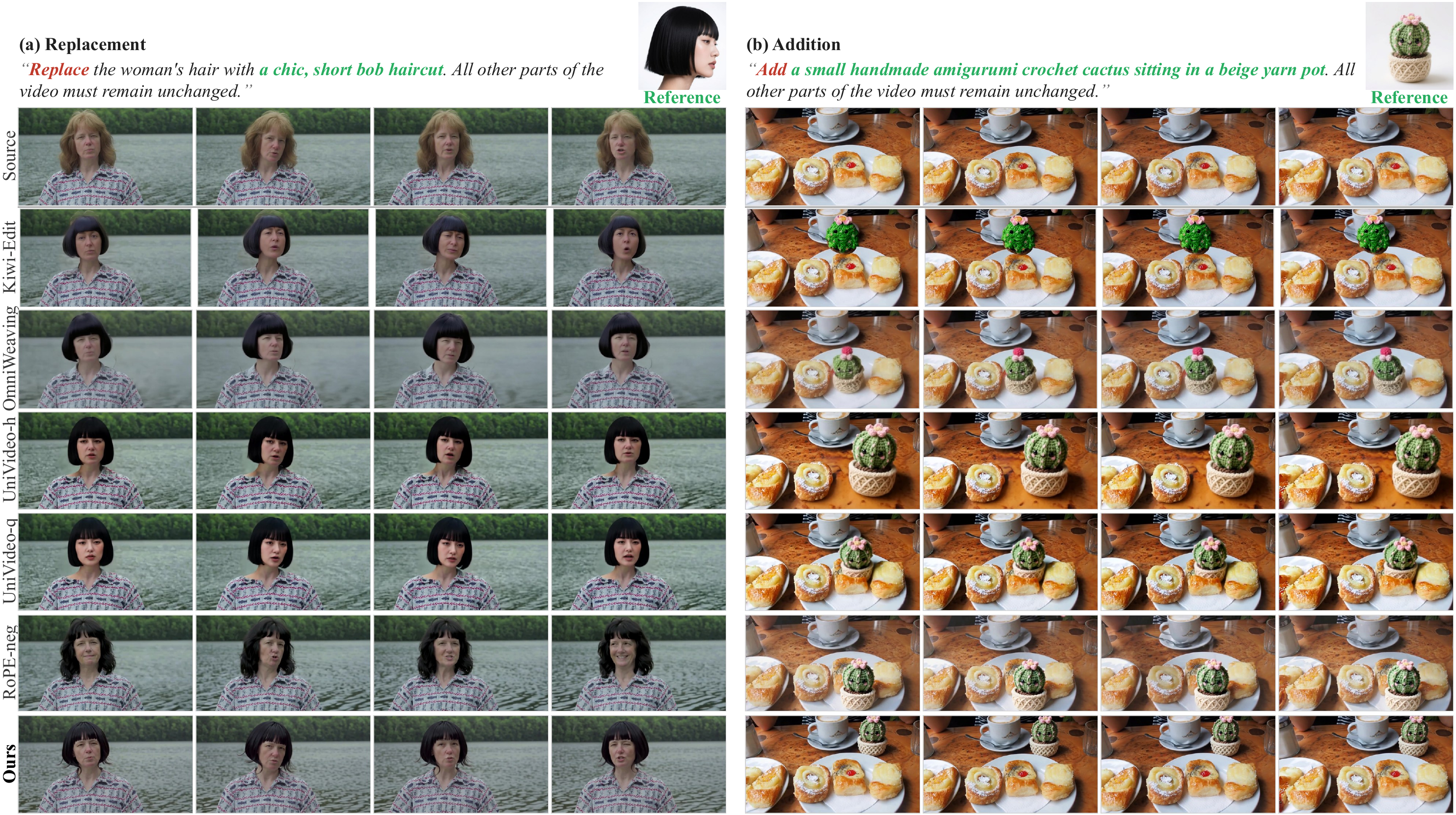}
\caption{\textbf{Qualitative comparison on RefVIE-Bench.} Two cases, subject replacement (left) and reference-guided addition (right), each shown as four frames sampled uniformly across the clip. The reference image is inset next to the instruction. Rows are ordered Source, open-source baselines, the \emph{w/ RoPE-Neg} variant, and TIDE.}
\label{fig:refvie_qualitative}
\end{figure*}

\noindent\textbf{Results on IntelligentVBench.}
To assess multimodal conditioning and composition, we report results on the TIV2V and Compositional MI2V tasks of IntelligentVBench in Tables~\ref{tab:tiv2v} and~\ref{tab:mi2v}.
On TIV2V, TIDE achieves the best Instruction Following, Condition Preserving, and average scores, improving the strongest baseline average from 3.89 to 4.10.
Although RoPE-Pos obtains a slightly higher Visual Quality score (3.80 versus 3.78), TIDE provides the strongest overall balance between faithfully executing the requested edit, preserving the reference identity and unedited source content, and maintaining visual quality.

\begin{table}[h]
\centering
\caption{Comparison on the TIV2V task of IntelligentVBench, under the official protocol (official prompts, Gemini-2.5-Pro judge). The \textbf{best} and \underline{second-best} scores in each column are marked.}
\label{tab:tiv2v}
\begin{tabular*}{\textwidth}{@{\extracolsep{\fill}}lcccc@{}}
\toprule
\textbf{Model} & \textbf{IF$\uparrow$} & \textbf{CP$\uparrow$} & \textbf{VQ$\uparrow$} & \textbf{AVG$\uparrow$} \\
\midrule
VACE-Wan2.1~\cite{jiang2025vace} & 1.46 & 1.42 & 1.71 & 1.53 \\
VACE-LTX~\cite{jiang2025vace} & 1.43 & 1.36 & 1.25 & 1.35 \\
VINO~\cite{chen2026vino} & 2.86 & 2.90 & 2.52 & 2.76 \\
UniVideo (query)~\cite{wei2025univideo} & 3.22 & 3.91 & 3.26 & 3.46 \\
UniVideo (hidden)~\cite{wei2025univideo} & 3.13 & 4.01 & 2.93 & 3.36 \\
OmniWeaving~\cite{pan2026omniweaving} & 4.00 & 4.04 & 3.65 & 3.89 \\
\midrule
TIDE w/ RoPE-Neg & 4.08 & 3.89 & 3.57 & 3.85 \\
TIDE w/ RoPE-Pos & \underline{4.09} & \underline{4.13} & \textbf{3.80} & \underline{4.01} \\
\textbf{TIDE (Ours)} & \textbf{4.21} & \textbf{4.29} & \underline{3.78} & \textbf{4.10} \\
\bottomrule
\end{tabular*}
\end{table}

\begin{figure*}[t]
\centering
\includegraphics[width=\textwidth]{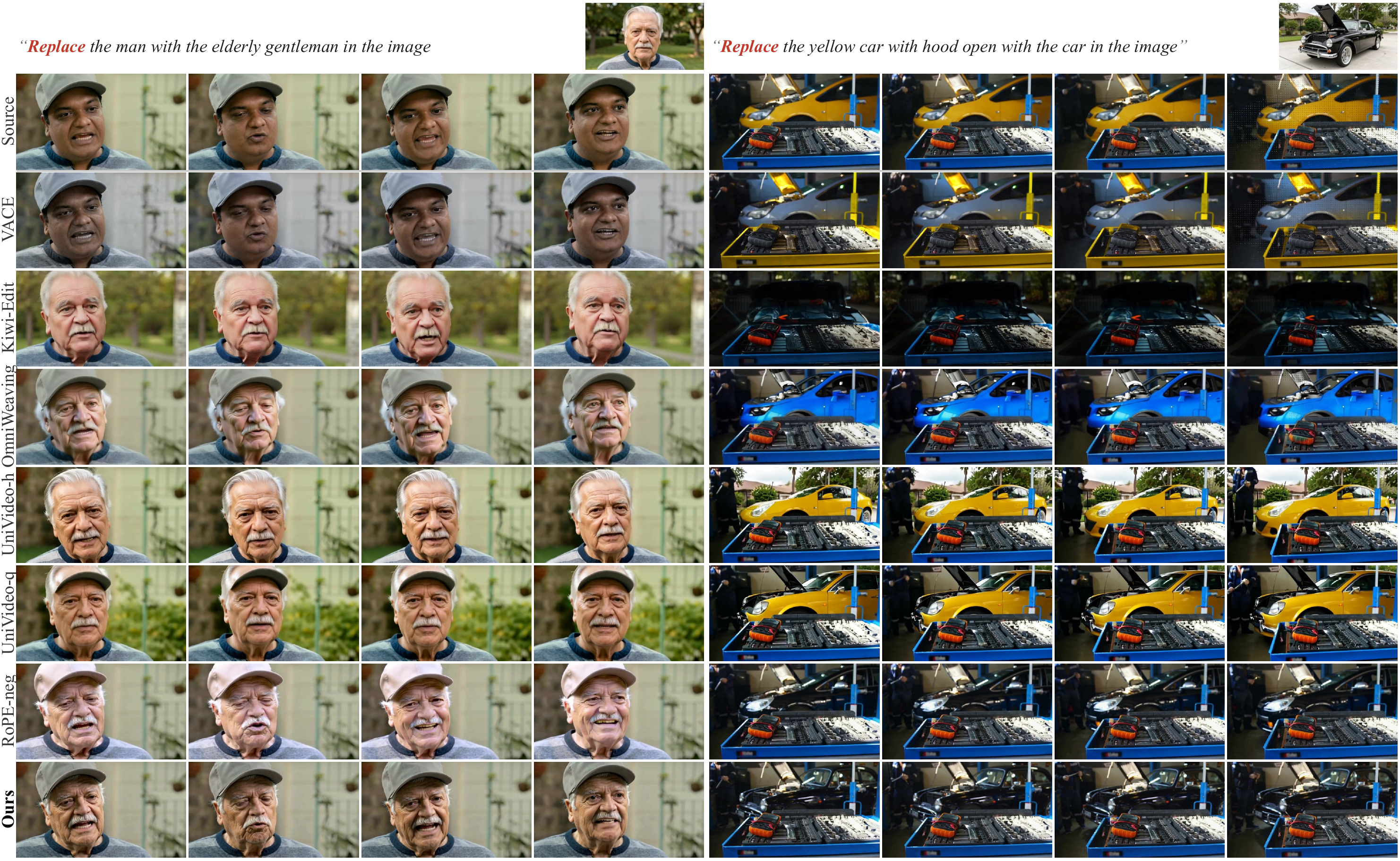}
\caption{\textbf{Qualitative comparison on the TIV2V task of IntelligentVBench.} Two local-change cases, each shown as four frames sampled uniformly across the clip, with the source video in the top row and the reference image inset next to the instruction. Baselines that alter unmentioned regions are visible as changes outside the target object.}
\label{fig:tiv2v_qualitative}
\end{figure*}

For the more challenging Compositional MI2V task, TIDE achieves the highest MIN scores for one- and two-subject generation and the highest average score in the two-subject setting, raising OmniWeaving's MIN and AVG from 3.61/4.27 to 4.01/4.39.
Despite not being trained on MI2V data, TIDE remains competitive in the one- and three-subject settings, demonstrating effective generalization from cropped multi-subject references to open-domain multi-image composition.
RoPE-Pos leads Instruction Following at every subject count and attains the best three-subject MIN, suggesting that the positive positional layout favors dense compositional prompting, while TIDE delivers more balanced reference preservation and generation quality in the one- and two-subject settings.

\begin{table}[h]
\centering
\caption{Comparison on the Compositional MI2V task of IntelligentVBench, under the official protocol. Sub-categories are defined by the number of reference subjects. MIN is the per-sample minimum over the three dimensions, averaged over samples. The \textbf{best} and \underline{second-best} scores in each column are marked.}
\label{tab:mi2v}
\resizebox{\textwidth}{!}{%
\setlength{\tabcolsep}{3pt}
\begin{tabular}{l|ccccc|ccccc|ccccc}
\toprule
\multirow{2}{*}{\textbf{Model}} & \multicolumn{5}{c|}{\textbf{1Subject (with BKG)}} & \multicolumn{5}{c|}{\textbf{2Subjects (with BKG)}} & \multicolumn{5}{c}{\textbf{3Subjects (with BKG)}} \\
 & IF$\uparrow$ & CP$\uparrow$ & VQ$\uparrow$ & MIN$\uparrow$ & AVG$\uparrow$ & IF$\uparrow$ & CP$\uparrow$ & VQ$\uparrow$ & MIN$\uparrow$ & AVG$\uparrow$ & IF$\uparrow$ & CP$\uparrow$ & VQ$\uparrow$ & MIN$\uparrow$ & AVG$\uparrow$ \\
\midrule
\multicolumn{16}{l}{\textit{Specialized Video Generation Models}} \\
SkyReels-A2~\cite{fei2025skyreelsa2} & 3.51 & 4.08 & 4.46 & 3.24 & 4.02 & 3.22 & 3.76 & 4.37 & 2.97 & 3.78 & 1.64 & 1.76 & 2.50 & 1.56 & 1.97 \\
SkyReels-V3~\cite{chen2025skyreels} & 3.46 & 3.71 & \underline{4.65} & 2.98 & 3.94 & 3.28 & 3.84 & 4.44 & 3.04 & 3.86 & 2.59 & 3.10 & 4.30 & 2.37 & 3.33 \\
MAGREF~\cite{deng2025magref} & 3.15 & 2.48 & 4.32 & 2.18 & 3.32 & 3.04 & 2.81 & 4.33 & 2.44 & 3.39 & 2.50 & 2.21 & \underline{4.46} & 2.07 & 3.06 \\
Phantom~\cite{liu2025phantom} & 3.21 & 2.95 & 4.29 & 2.47 & 3.48 & 2.88 & 3.42 & 4.38 & 2.62 & 3.55 & 2.36 & 2.79 & 4.21 & 2.20 & 3.12 \\
\midrule
\multicolumn{16}{l}{\textit{Unified Video Generation Models}} \\
VACE-Wan2.1~\cite{jiang2025vace} & 3.88 & \underline{4.48} & \textbf{4.68} & 3.62 & 4.35 & 3.31 & 4.03 & 4.51 & 3.08 & 3.95 & 2.60 & 3.03 & 4.40 & 2.50 & 3.34 \\
VACE-LTX~\cite{jiang2025vace} & 2.74 & 2.86 & 2.89 & 2.25 & 2.83 & 2.12 & 2.26 & 2.49 & 1.98 & 2.29 & 1.94 & 2.06 & 2.41 & 1.89 & 2.14 \\
VINO~\cite{chen2026vino} & 3.72 & 4.22 & 4.46 & 3.42 & 4.13 & 3.56 & \underline{4.34} & \textbf{4.58} & 3.39 & 4.16 & 2.63 & 2.97 & 4.24 & 2.53 & 3.28 \\
UniVideo (query)~\cite{wei2025univideo} & 3.35 & 3.90 & 4.41 & 3.01 & 3.89 & 2.98 & 3.73 & 4.18 & 2.76 & 3.63 & 2.30 & 2.50 & 3.89 & 2.20 & 2.89 \\
UniVideo (hidden)~\cite{wei2025univideo} & 3.33 & 4.18 & 4.38 & 3.08 & 3.97 & 3.22 & 4.12 & 4.36 & 3.09 & 3.90 & 2.31 & 2.83 & 3.94 & 2.29 & 3.03 \\
OmniWeaving~\cite{pan2026omniweaving} & \underline{4.35} & \textbf{4.53} & 4.58 & \underline{4.01} & \textbf{4.49} & 4.08 & 4.22 & \underline{4.52} & 3.61 & 4.27 & 3.53 & \textbf{4.01} & \textbf{4.54} & 3.26 & \textbf{4.03} \\
\midrule
TIDE w/ RoPE-Neg & 4.07 & 4.23 & 4.25 & 3.74 & 4.18 & 4.10 & 4.24 & 4.32 & \underline{3.86} & 4.22 & 3.63 & 3.53 & 3.96 & 3.17 & 3.70 \\
TIDE w/ RoPE-Pos & \textbf{4.86} & 4.25 & 4.05 & 3.78 & 4.39 & \textbf{4.52} & 4.15 & 4.20 & 3.75 & \underline{4.29} & \textbf{4.30} & 3.77 & 3.91 & \textbf{3.46} & \underline{4.00} \\
\textbf{TIDE (Ours)} & 4.28 & 4.37 & 4.54 & \textbf{4.03} & \underline{4.40} & \underline{4.23} & \textbf{4.37} & \textbf{4.58} & \textbf{4.01} & \textbf{4.39} & \underline{3.67} & \underline{3.83} & 4.27 & \underline{3.41} & 3.92 \\
\bottomrule
\end{tabular}
}
\end{table}

\begin{figure*}[t]
\centering
\includegraphics[width=\textwidth]{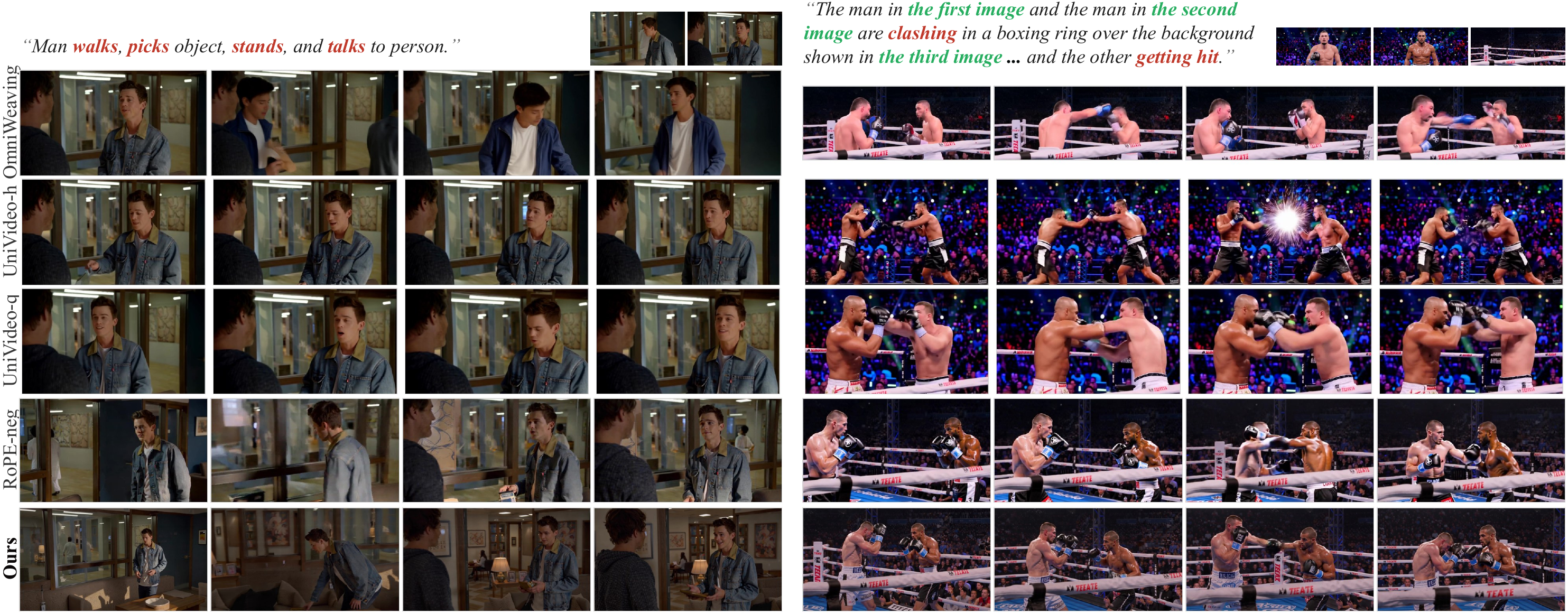}
\caption{\textbf{Qualitative comparison on the Compositional MI2V task of IntelligentVBench.} Two multi-subject cases, each shown as four frames sampled uniformly across the clip; the reference images are inset next to the prompt. There is no source video for this task, so all rows are generated results. Baselines merge or swap subject identities as the reference count grows, while TIDE keeps each reference bound to its own subject.}
\label{fig:mi2v_qualitative}
\end{figure*}

\subsection{Qualitative Results}
\label{sec:qualitative}

As illustrated in Figures~\ref{fig:qualitative_edit}--\ref{fig:mi2v_qualitative}, we qualitatively compare TIDE with the baselines available for each benchmark.
On OpenVE-Bench, TIDE follows the editing instructions more accurately than Ditto and VINO while better preserving unedited regions.
For reference-guided editing on RefVIE-Bench and TIV2V, TIDE more faithfully restores the supplied subject or background reference, applies it to the intended target, and produces higher-fidelity editing results.
On TIDE-Bench, TIDE accurately binds each reference image to the corresponding edit and preserves unrelated scene content, including the tabletop in the second case.
For multi-reference generation on OpenS2V and Compositional MI2V, TIDE better maintains the distinct identity of every reference subject while following the requested interactions and retaining favorable overall video quality.
Across all displayed cases, TIDE provides the most faithful and visually coherent result; additional qualitative examples are included in the supplementary material.

\subsection{Ablation Studies}
\label{sec:ablation}

We ablate TIDE's core design choices on TIDE-Bench and OpenVE-Bench.

\noindent\textbf{Task Embedding.}
We compare against two positional alternatives throughout the benchmark tables: RoPE-Neg places conditioning tokens before the target on a negative position axis, while RoPE-Pos keeps the target starting at zero and appends conditioning tokens on the positive axis.
The final task-embedding model surpasses both variants on OpenVE-Bench (2.91 vs.\ 2.37/2.24) and TIDE-Bench (3.41 vs.\ 2.97/2.66).
Its Reference Faithfulness score also improves from 3.00/2.71 to 3.44, supporting explicit per-token task identifiers for distinguishing visual conditions.

\begin{wraptable}[9]{r}{0.41\textwidth}
\vspace{-0.8\baselineskip}
\centering
\captionsetup{font=small,skip=3pt}
\caption{Ablation study on components and progressive training.}
\label{tab:ablation_method}
\footnotesize
\setlength{\tabcolsep}{4pt}
\begin{tabular}{lcc}
\toprule
\textbf{Method} & \textbf{TIDE$\uparrow$} & \textbf{OpenVE$\uparrow$} \\
\midrule
w/o VLM & 2.56 & 1.98 \\
w/o VLM-Source & 2.89 & 2.01 \\
\midrule
Stage 1 & 2.51 & 2.99 \\
Stage 2 & 3.30 & 2.84 \\
\midrule
\rowcolor{gray!10}
\textbf{TIDE (Stage 3)} & \textbf{3.41} & \textbf{2.91} \\
\bottomrule
\end{tabular}
\vspace{-0.4\baselineskip}
\end{wraptable}

\noindent\textbf{Component Ablation.}
Table~\ref{tab:ablation_method} (top) shows training-matched variants.
Removing the VLM path causes the largest drop on both benchmarks.
Encoding references but not the source video recovers part of the performance, indicating that source-side VLM features provide additional guidance.

\noindent\textbf{Progressive Training.}
Table~\ref{tab:ablation_method} (bottom) tracks per-stage performance.
Stage~2 introduces multi-task data and yields a large gain on TIDE-Bench ($+$0.79), while Stage~3 adds a further $+$0.11.
On OpenVE-Bench, Stage~1 scores 2.99, while Stages~2 and~3 trade a small decrease for improved multi-reference performance.

We report the CFG-scale sweep in the supplementary material.

\section{Conclusion}
\label{sec:conclusion}

We presented TIDE, a unified framework that integrates instruction-based video editing, reference-guided video editing, and multi-reference video generation through a joint Diffusion Transformer and vision-language model architecture, equipped with per-token task embeddings and dual-path conditioning.
Experiments on OpenVE-Bench, TIDE-Bench, OpenS2V, RefVIE-Bench, and the TIV2V and Compositional MI2V tasks of IntelligentVBench demonstrate state-of-the-art or competitive performance across both editing and generation tasks.

\noindent\textbf{Limitations.}
Despite strong overall performance, several limitations remain.
First, our training data lacks multi-shot editing sequences, limiting TIDE's ability to perform coherent edits across scene transitions and camera cuts; collecting or synthesizing multi-shot editing data would strengthen this capability.
Second, TIDE currently operates exclusively in the visual domain and does not support audio-aware editing (e.g., synchronizing speech or sound effects with visual modifications), an increasingly important direction as audio-visual generation models mature~\cite{hacohen2026ltx,low2025ovi}.
Future work will explore extending the task embedding framework to accommodate multi-shot and audio-visual editing scenarios.

\FloatBarrier

\bibliographystyle{Ref}
\bibliography{Ref}

\newpage
\appendix

\begin{center}
  {\LARGE\bfseries Appendix / Supplementary Material}\par
  \vspace{0.35em}
  {\large \emph{TIDE}: Task-Isolated Diffusion for Unified Video Editing and Generation}
\end{center}
\vspace{0.75em}

\begin{tcolorbox}[
  enhanced,
  colback=GoogleBlue!3,
  colframe=GoogleBlue!55!black,
  colbacktitle=GoogleBlue!10,
  coltitle=black,
  title={\large\bfseries Appendix Contents},
  boxrule=0.7pt,
  arc=2mm,
  left=3mm,right=3mm,top=2mm,bottom=2mm
]
\small
\appcontentsline{app:ablation_qualitative}{Qualitative Ablation Analysis}
\appcontentsline{app:cfg_ablation}{Guidance Scale}
\appcontentsline{app:tidebench}{TIDE-Bench Details}
\appcontentssubline{app:tidebench_source}{Data Source and Construction}
\appcontentssubline{app:tidebench_protocol}{Evaluation Protocol}
\appcontentssubline{app:tidebench_prompts}{Judge Prompts}
\appcontentsline{app:training_data}{Training Data and Stage Configuration}
\appcontentsline{app:pipeline_edit}{Reference-Guided Editing Data Construction}
\appcontentsline{app:pipeline_ip}{Subject-to-Video Data Construction}
\appcontentsline{app:additional_qualitative}{Additional Qualitative Results}
\appcontentssubline{app:tidebench_extended}{Extended Comparisons on TIDE-Bench}
\appcontentssubline{app:multiref_cases}{Multi-Reference Video Editing Cases}
\appcontentssubline{app:s2v_cases}{Subject-to-Video Generation Cases}
\end{tcolorbox}

\clearpage

\section{Qualitative Ablation Analysis}
\label{app:ablation_qualitative}

\begin{figure}[h]
\centering
\includegraphics[width=\textwidth]{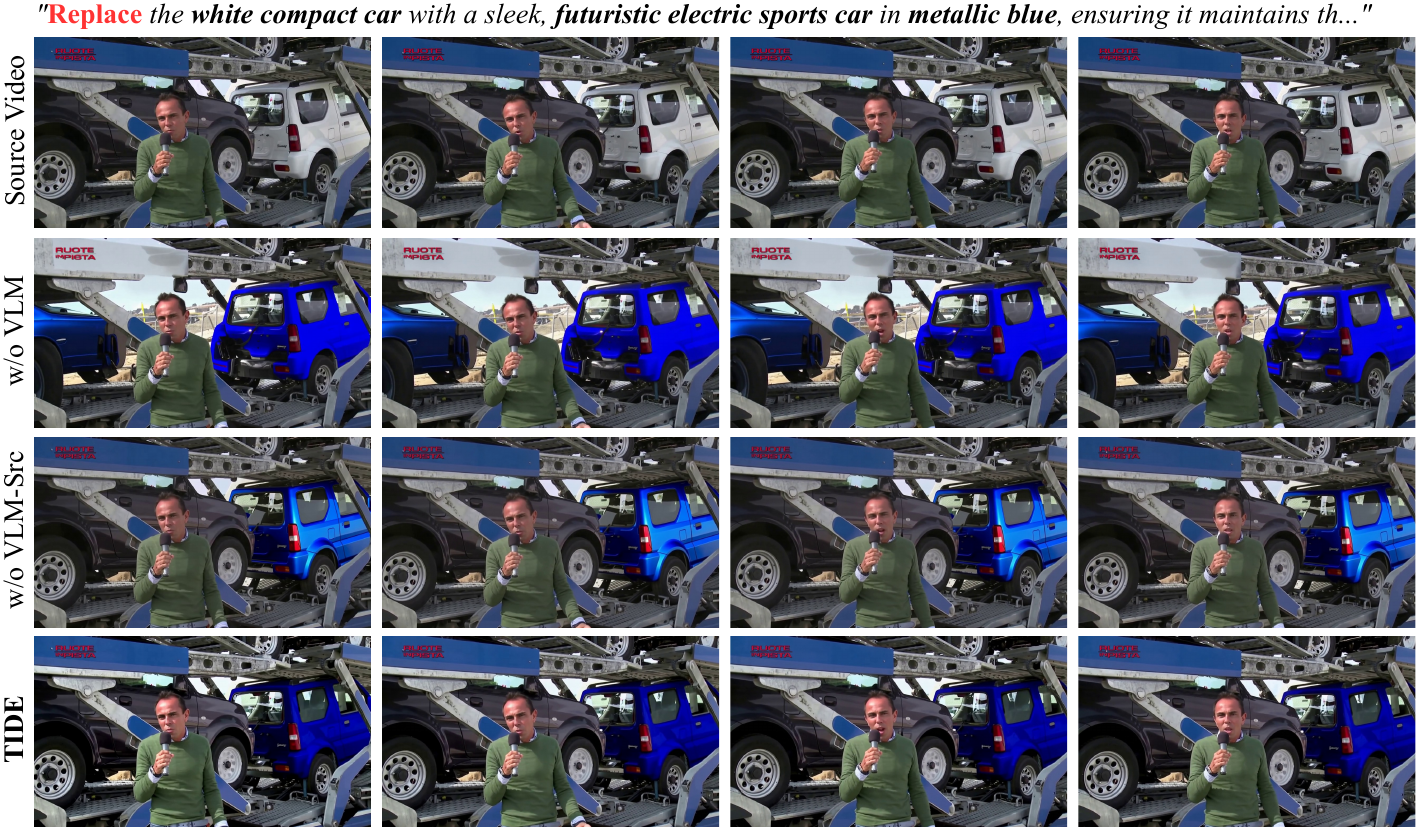}
\caption{Qualitative ablation on an OpenVE-Bench editing case. Only the full TIDE model faithfully follows the instruction while preserving the scene.}
\label{fig:ablation_qualitative}
\end{figure}

Figure~\ref{fig:ablation_qualitative} presents a qualitative comparison of ablation variants on an OpenVE-Bench editing case, where the instruction asks to replace a white compact car with a metallic blue electric sports car.
\emph{w/o VLM} removes the original black car from the scene entirely rather than performing the intended replacement, failing to interpret the editing instruction correctly.
\emph{w/o VLM-Source} successfully replaces the car with a blue vehicle, but the left rear wheel of the replacement car is missing, indicating degraded structural fidelity without VLM-based source scene understanding.
Only the full TIDE model correctly generates a complete blue sports car while maintaining temporal consistency and preserving the reporter and background scene.

\begin{figure}[h]
\centering
\includegraphics[width=\textwidth]{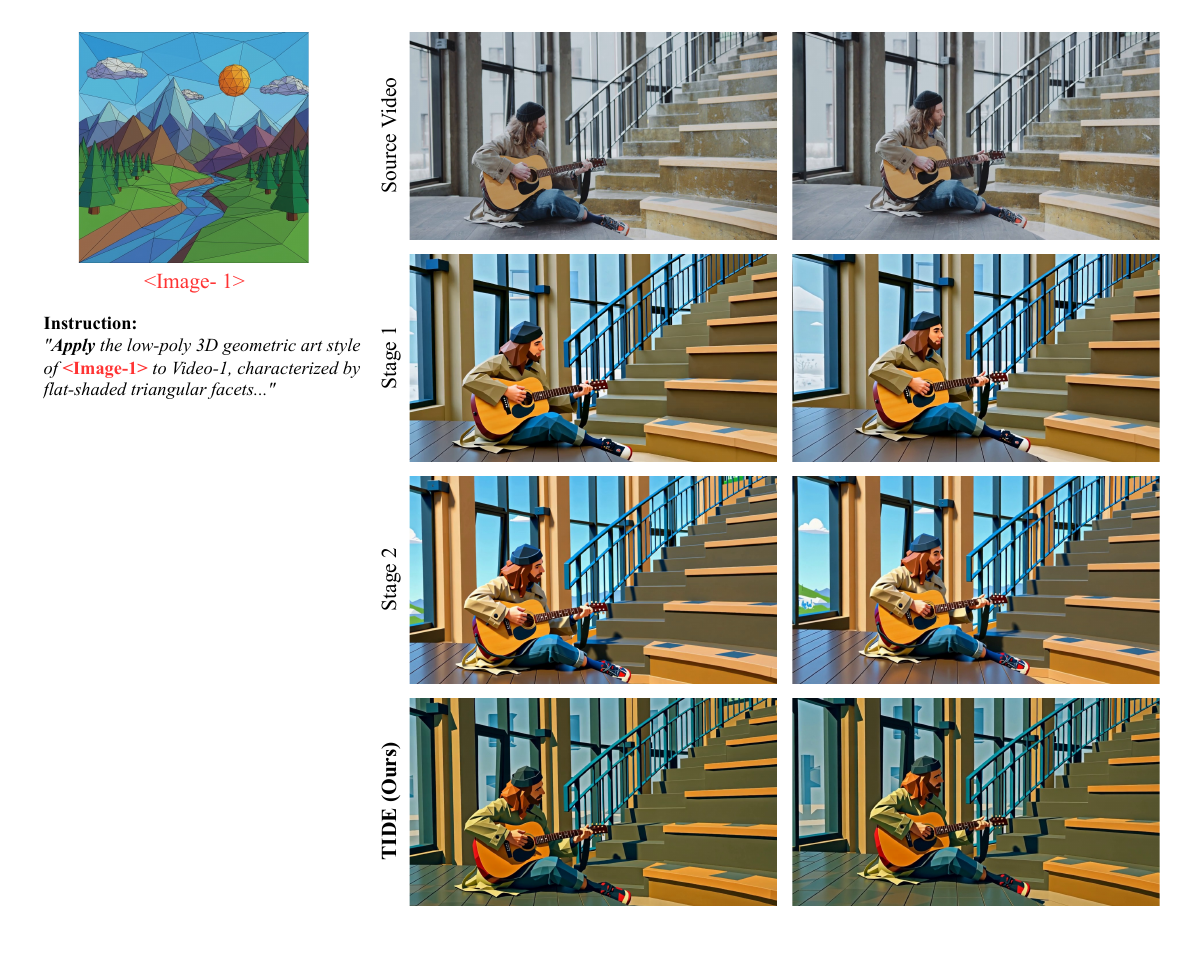}
\caption{Progressive training ablation on a TIDE-Bench style transfer case. Stage~1 conflates reference content with style; Stage~2 partially leaks; TIDE (Full) transfers only the style.}
\label{fig:ablation_stages}
\end{figure}

Figure~\ref{fig:ablation_stages} shows a progressive training ablation on a TIDE-Bench style transfer case, where the instruction asks to apply a low-poly 3D geometric art style from a reference image to a source video of a person playing guitar on stairs.
Stage~1 (instruction-based video editing only) applies the low-poly style but also introduces content from the reference image (mountains, river) into the background, failing to disentangle style from content.
Stage~2 improves content preservation (the person and guitar are maintained), but the style transfer leaks reference content into the background (cartoon-like sky and buildings visible through the window), and overall stylization is inconsistent.
The full TIDE model successfully transfers only the low-poly geometric art style while faithfully preserving the original video content: the person, guitar, stairs, and indoor setting remain intact with consistent stylization across frames, demonstrating that progressive multi-task training is critical for learning to separate style from content in reference-guided editing.

\section{Guidance Scale}
\label{app:cfg_ablation}

\begin{figure}[h]
\centering
\includegraphics[width=0.48\textwidth]{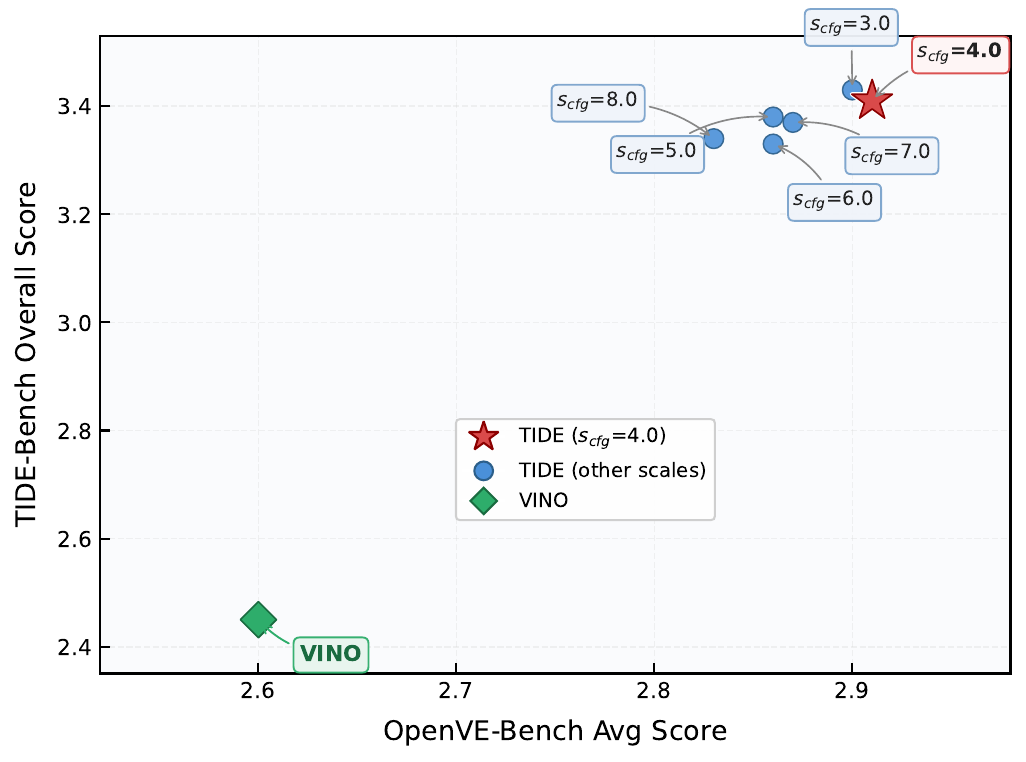}
\caption{Effect of CFG scale $s_{\text{cfg}}$ on OpenVE-Bench and TIDE-Bench. Star marks $s_{\text{cfg}}{=}4.0$.}
\label{fig:cfg_ablation}
\end{figure}

Figure~\ref{fig:cfg_ablation} shows the effect of the CFG scale $s_{\text{cfg}}$ on OpenVE-Bench and TIDE-Bench.
Performance is robust across $s_{\text{cfg}} \in [3.0, 8.0]$, with $s_{\text{cfg}}{=}4.0$ achieving the best overall trade-off: it attains the highest OpenVE-Bench score (2.91) while maintaining strong TIDE-Bench performance (3.41).
Lower scales ($s_{\text{cfg}}{=}3.0$) slightly favor multi-reference editing at the expense of instruction compliance, while higher scales ($s_{\text{cfg}} \geq 6.0$) degrade both metrics.
We therefore adopt $s_{\text{cfg}}{=}4.0$ for all main experiments.

\section{TIDE-Bench Details}
\label{app:tidebench}

\subsection{Data Source and Construction}
\label{app:tidebench_source}

TIDE-Bench is designed to test whether a video editor can jointly follow compound instructions, bind each visual condition to the correct edit, and preserve the remainder of the source video.
We collect 56 source videos from the Pexels website~\citep{pexels} and filter them for scene diversity, visible motion, and sufficient content for localized or compositional editing.
From these videos, we construct 211 editing scenarios.
Each scenario contains a source video, an editing instruction, up to three reference images, and explicit metadata specifying the role of each reference.

\subsection{Evaluation Protocol}
\label{app:tidebench_protocol}

We adopt the LLM-as-judge paradigm and use Gemini-3.1-Pro~\cite{gemini31pro} as the automated evaluator.
For every scenario, the judge receives the visual inputs in a fixed order: the original video, all applicable reference images in their metadata order, and the edited video.
The instruction and a role description for each reference are inserted into the prompt.
We use separate complete prompt templates for global style transfer and local or compound editing, as provided in Appendix~\ref{app:tidebench_prompts}.

Each output is scored on five 1\,\textendash\,5 dimensions: \emph{Edit Completeness}, \emph{Reference Faithfulness}, \emph{Visual \& Temporal Quality}, \emph{Scene Preservation}, and \emph{Overall Quality}.
For a scenario with multiple references, the judge scores every reference independently and Reference Faithfulness is the minimum of these per-reference scores, thereby exposing failure on any individual condition rather than allowing it to be hidden by averaging.
Reference Faithfulness is averaged over applicable scenarios, while the other four dimensions are averaged over all evaluated scenarios.
The reported composite score is the unweighted mean of the five resulting dimension-level means.

To prevent inflated scores from naive copy-pasting, Reference Faithfulness, Visual \& Temporal Quality, Scene Preservation, and Overall Quality are constrained to be no greater than Edit Completeness.
The prompt further directs the judge to penalize reference content that appears as a crude overlay without plausible integration.
Together, the worst-reference aggregation and score cap require a method to execute every requested operation before it can receive a high composite score.

\subsection{Judge Prompts}
\label{app:tidebench_prompts}

For reproducibility, we provide the complete judge templates below.
At evaluation time, the four placeholders are populated as follows:
\begin{itemize}[leftmargin=1.5em,itemsep=0pt,topsep=2pt]
  \item \texttt{<EDITING\_INSTRUCTION>}: the scenario instruction;
  \item \texttt{<REFERENCE\_ROLES>}: each reference and its intended role;
  \item \texttt{<PER\_REFERENCE\_CHECKLIST>}: the element associated with each reference; and
  \item \texttt{<EDIT\_TYPE>}: the applicable operation combination.
\end{itemize}
These fields are generated directly from each scenario's metadata; the scoring criteria and output schema remain fixed.

\begin{judgeprompt}{Judge prompt for global reference-guided style transfer}
You are an expert evaluator for video style transfer editing. You will be provided with three visual inputs in order:
1. **[Original Video]**: The source video before any editing.
2. **[Reference Image-1]**: A style reference image that defines the target artistic style.
3. **[Edited Video]**: The result after applying the style transfer.

Your task is to carefully compare all three and evaluate how well the style transfer was performed.

**Editing instruction**: <EDITING_INSTRUCTION>

<REFERENCE_ROLES>

Evaluate on five dimensions, each on a 1-5 integer scale:

**Edit Completeness (Style Fidelity)**
Assess whether the target artistic style has been fully and consistently applied across the entire video duration.
5 = Full, faithful transfer: colour palette, texture, brushwork/artistic techniques match the style reference consistently over every frame of the entire video
4 = Style reproduced well across almost the whole video; only small local regions or brief temporal moments show mismatches
3 = Key style traits (e.g., colour palette or texture) are present but application is patchy, inconsistent across frames, or fades in/out
2 = Style shows in only a few areas or frames; rest of the video retains original appearance or shows unrelated styles
1 = Target style is absent, barely detectable, or a completely wrong style was applied

**Reference Faithfulness**
<PER_REFERENCE_CHECKLIST>
Score EACH reference image independently on a 1-5 scale. Compare the edited video's visual style against the reference image:
5 = Style characteristics (colour palette, texture patterns, brushwork, artistic technique) exactly match the reference image throughout the video
4 = Strong resemblance to reference style; the artistic approach is clearly the same, only fine details or subtle tonal differences
3 = General style direction is correct (e.g., right artistic movement) but noticeable differences in palette, texture density, or technique
2 = Vaguely related style but substantially different from the reference -- wrong colour scheme, different texture, or inconsistent technique
1 = No discernible resemblance to the reference style image at all

**Visual & Temporal Quality**
Assess the visual fidelity and temporal stability of the edited video. Pay attention to flickering, boiling textures, edge artifacts, and frame-to-frame consistency.
5 = Perfectly stable and temporally coherent; no flickering, clean edges, high resolution maintained throughout
4 = Largely stable with only minor, subtle flickering in areas of complex motion; quality is high
3 = Noticeable but tolerable flicker or texture "boiling", especially during fast motion or scene transitions
2 = Significant and distracting flickering, jittering, or temporal inconsistency that degrades viewing experience
1 = Extreme flickering or "boiling" effects; video is essentially unwatchable

**Scene Preservation (Content Preservation)**
Assess whether the original video's content -- objects, spatial layout, motion trajectories, perspective -- is preserved. Only stylistic changes should be made.
5 = All objects, spatial relations, and motion perfectly kept; only stylistic changes applied
4 = Nearly all geometry and motion intact; only slight, non-distracting deformation in minor areas
3 = Overall structure and motion direction correct; some local warping, slight motion jerkiness, or minor spatial distortion
2 = Main subject recognisable, but size, perspective, motion speed, or key body parts are clearly wrong
1 = Major objects, layout, or overall motion lost/distorted; scene is barely recognisable as the original

**Overall Quality**
Holistic assessment of the final video's quality as a creative output.
5 = Publication-ready; aesthetically pleasing, high resolution, physically plausible, no visible artifacts
4 = Good quality; minor imperfections only visible on close inspection
3 = Acceptable quality; some noticeable artifacts or quality issues but still viewable
2 = Poor quality; obvious artifacts, blur, or distortion that significantly degrade the result
1 = Very poor; heavily degraded, unwatchable

RULES: Reference Faithfulness, Visual & Temporal Quality, Scene Preservation, and Overall Quality scores must ALL be <= Edit Completeness score. If the reference content appears as a crude overlay or copy-paste with poor scene integration, the Reference Faithfulness score must be penalised accordingly (cap at Edit Completeness score).

Respond in EXACTLY this format (nothing else):
Brief reasoning: <one sentence, max 30 words>
Edit Completeness: <1-5>
Reference Faithfulness: Ref-1=<1-5> (one score per reference image)
Visual & Temporal Quality: <1-5>
Scene Preservation: <1-5>
Overall Quality: <1-5>
\end{judgeprompt}

\begin{judgeprompt}{Judge prompt for local and compound video editing}
You are an expert evaluator for multi-reference video editing. You will be provided with visual inputs in order:
1. **[Original Video]**: The source video before any editing.
2. **[Reference Image-1], [Reference Image-2], ...**: Reference images that define what should be added, replaced, or used as background (if applicable).
3. **[Edited Video]**: The result after applying the editing operations.

Your task is to carefully compare all inputs and evaluate how well the editing was performed according to the instruction.

**Editing instruction**: <EDITING_INSTRUCTION>

<REFERENCE_ROLES>

**Edit type**: <EDIT_TYPE>

Evaluate on five dimensions, each on a 1-5 integer scale. Use the dimension name (not D1/D2/...) as the section header.

**Edit Completeness (Prompt Compliance)**
Assess whether ALL requested editing operations were actually performed correctly and consistently throughout the video duration. Consider both whether each operation happened and whether it targeted the correct subjects.
5 = Every requested operation fully executed with correct targets, maintained consistently for the entire video duration
4 = All operations attempted and mostly successful; one minor deficiency (e.g., slight inconsistency in a few frames)
3 = At least one operation is only partially done, inconsistently applied, or missing from some frames
2 = Multiple operations missing, only superficially executed, or applied to wrong targets
1 = No discernible editing performed, or an entirely wrong/unrelated edit was applied

**Reference Faithfulness**
<PER_REFERENCE_CHECKLIST>
Score EACH reference image independently on a 1-5 scale. For each reference, assess how faithfully its visual content (shape, colour, texture, identity-defining features) was reproduced in the edited video:
5 = Shape, colour, texture, and distinctive features exactly match the reference; the element is immediately recognisable as the same object/scene
4 = Strong resemblance; correct category and key attributes preserved; only fine details (e.g., minor texture, exact proportions) differ
3 = Correct category but noticeable drift -- wrong shade/colour, simplified geometry, missing distinctive features
2 = A vaguely related element appears but is substantially different from the reference in shape, colour, or identity
1 = Referenced content is completely absent from the edited video, or an entirely unrelated element was used

**Visual & Temporal Quality**
Assess the visual fidelity and temporal stability of the edited video. Focus on the edit regions but also check for artifacts introduced elsewhere. Pay attention to flickering, edge seams, colour bleeding, and frame-to-frame consistency.
5 = Perfectly seamless integration; edit area is undetectable, no flicker, jitter, or quality degradation anywhere
4 = Style and quality almost uniform and stable; tiny temporal artefacts visible only on frame-by-frame inspection
3 = Basic quality acceptable but noticeable issues: lighting/palette clashes, inconsistent across frames, minor flickering in edit regions
2 = Obvious seams/edges around edited areas, strong colour mismatch with surroundings, significant flickering or jittering
1 = Video heavily broken, massive distortion, or uncontrollable flicker; edit regions are clearly artificial

**Scene Preservation (Physical Coherence)**
Assess whether regions that should NOT be edited remain unchanged, and whether the overall physical plausibility (motion, lighting, shadows, perspective) is maintained.
5 = Unedited regions are pixel-faithful to original; motion, lighting, shadows, perspective all physically correct and temporally stable
4 = Visually identical to original under casual viewing; only subtle deformation detectable on close inspection in non-edit areas
3 = Most of the scene preserved but noticeable artifacts in untargeted areas; minor physics errors (wrong shadows, slight perspective shift)
2 = Significant unintended changes: subject deformation, objects vanishing, wrong shadows, or spatial layout altered
1 = Original content largely destroyed or unrecognisable; massive unintended modifications throughout

**Overall Quality**
Holistic assessment of the final video as a creative output, considering aesthetic appeal, resolution, and freedom from artifacts.
5 = Publication-ready; aesthetically pleasing, high resolution, physically plausible, no visible artifacts
4 = Good quality overall; minor imperfections only visible on close inspection
3 = Acceptable quality; some noticeable artifacts, blur, or quality issues but still viewable
2 = Poor quality; obvious artifacts, distortion, or degradation that significantly degrade the result
1 = Very poor; heavily degraded, unwatchable

RULES: Reference Faithfulness, Visual & Temporal Quality, Scene Preservation, and Overall Quality scores must ALL be <= Edit Completeness score. If the reference content appears as a crude overlay or copy-paste with poor scene integration, the Reference Faithfulness score must be penalised accordingly (cap at Edit Completeness score). If no reference images are provided, skip Reference Faithfulness and write "Reference Faithfulness: N/A".

Respond in EXACTLY this format (nothing else):
Brief reasoning: <one sentence, max 30 words>
Edit Completeness: <1-5>
Reference Faithfulness: Ref-1=<1-5>, Ref-2=<1-5>, ... (one score per reference image; if only 1 ref, just Ref-1=<1-5>; if no refs, write N/A)
Visual & Temporal Quality: <1-5>
Scene Preservation: <1-5>
Overall Quality: <1-5>
\end{judgeprompt}

\section{Training Data and Stage Configuration}
\label{app:training_data}

\noindent\textbf{Training Data.}
Our training pool comprises ${\sim}$1.83M samples organized into five categories:
(i)~\emph{Reference-guided video editing} (${\sim}$730K): ${\sim}$700K single-reference and ${\sim}$30K multi-reference editing samples covering addition, removal, replacement, and style-reference editing (construction details in Appendix~\ref{app:pipeline_edit});
(ii)~\emph{Instruction-based video editing} (${\sim}$500K): source-target video pairs with editing instructions from Ditto~\cite{ditto} and OpenVE~\cite{openve};
(iii)~\emph{Image editing} (${\sim}$170K): multi-reference image editing data from MICo~\cite{mico} and Unipic~\cite{unipic}, treated as single-frame video;
(iv)~\emph{Subject-to-video generation} (${\sim}$330K): text-subject-to-video data with cross-video identity pairing (Appendix~\ref{app:pipeline_ip});
(v)~\emph{Reference-guided style video editing} (${\sim}$100K): style-level reference-conditioned video editing pairs.

\begin{figure}[h]
\centering
\includegraphics[width=0.85\textwidth]{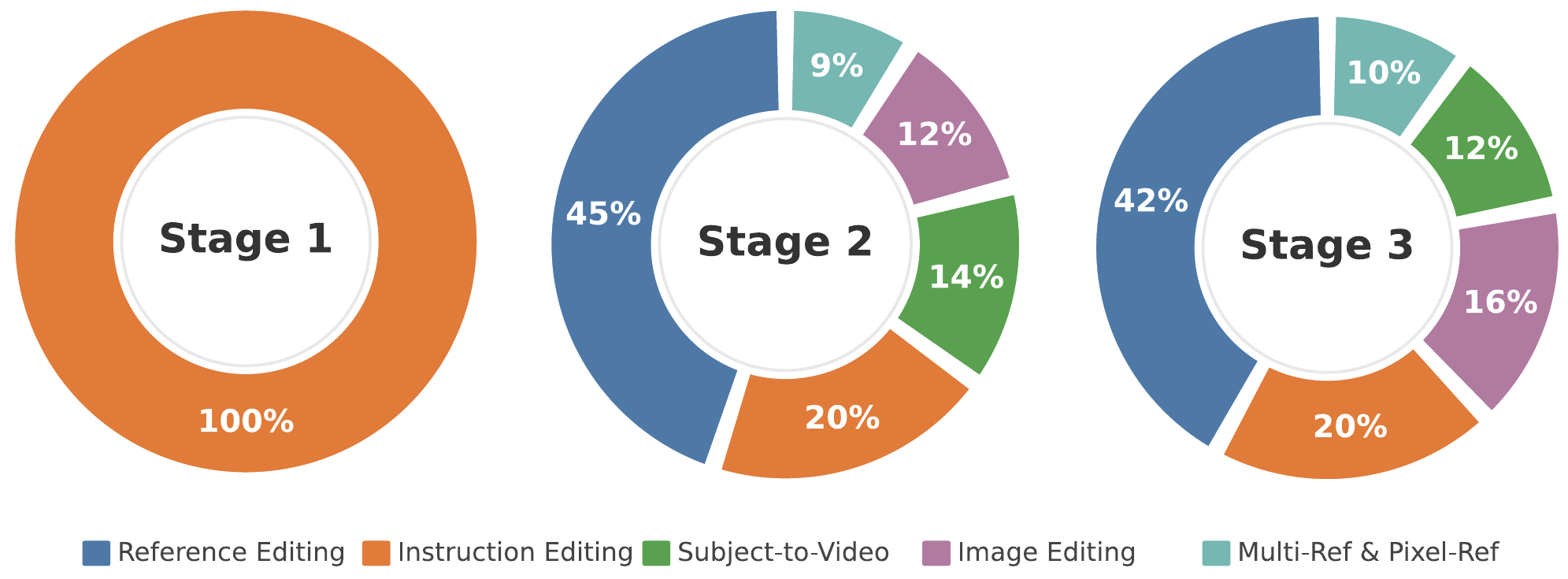}
\caption{Data composition ratios across the three progressive training stages. Stage~1 uses only instruction-based video editing data; Stage~2 introduces the full multi-task mixture; Stage~3 refines sampling ratios to balance underrepresented categories.}
\label{fig:data_pie}
\end{figure}

\noindent\textbf{Progressive Stage Configuration.}
Training proceeds through three stages of increasing task complexity.
\emph{Stage~1} (${\sim}$3K steps, lr $=1\times10^{-4}$) trains exclusively on instruction-based video editing~(ii) to establish basic instruction-following and video-editing competence before any reference conditioning is introduced.
\emph{Stage~2} (${\sim}$7K steps, lr $=5\times10^{-5}$) activates the full multi-task mixture, adding reference-guided video editing~(i), image editing~(iii), subject-to-video generation~(iv), and reference-guided style editing~(v), with per-category sampling ratios set proportionally to the available data.
\emph{Stage~3} (${\sim}$10K steps, lr $=2\times10^{-5}$) continues multi-task training with refined sampling ratios that up-weight the underrepresented reference-guided style~(v) and subject-to-video~(iv) categories, and additionally emphasize the multi-reference subset within~(i), for long-horizon convergence on the hardest scenarios.
All stages fine-tune the full DiT backbone together with the task embedding tables while keeping the VLM encoder frozen; step counts are normalized to 32~H20 GPUs with gradient accumulation of~4 (effective batch size~128).
The per-stage data composition ratios over the five major categories are illustrated in Figure~\ref{fig:data_pie}.

\section{Reference-Guided Editing Data Construction}
\label{app:pipeline_edit}

We construct large-scale reference-guided video editing data from existing open-source instruction-based editing datasets (OpenVE-3M~\cite{openve} and Ditto~\cite{ditto}) through an automated extraction-and-filtering pipeline.
Starting from the available (source video, target video, instruction) triplets, our goal is to obtain (source video, target video, reference image, instruction) quadruplets.
The pipeline operates in two phases: \emph{single-reference extraction} converts instruction-edit video pairs into reference-guided quadruplets, and \emph{multi-reference extension} constructs multi-reference training samples through iterative editing.

\noindent\textbf{Single-Reference Extraction.}
Given a (source video, target video, instruction) triplet from the instruction-based editing datasets, the pipeline proceeds through four stages:
\begin{enumerate}[leftmargin=*,nosep]
    \item \textbf{First-Frame Extraction.}
    We extract the first frame from both the source video and the target video, producing a source frame and a target frame that capture the visual difference introduced by the edit.

    \item \textbf{Reference Prompt Generation.}
    A VLM (Gemini~\cite{gemini31flashlite}) takes the source frame, target frame, and the original editing instruction as input to generate a descriptive prompt for extracting a clean reference image.
    The generated prompt describes the visual element that should be isolated as the reference (e.g., the added object, the replacement subject, or the target style).

    \item \textbf{Reference Image Extraction.}
    The source frame, target frame, and the generated reference prompt are fed into Qwen Image Edit~\cite{wu2025qwenimagetechnicalreport} to extract a clean, isolated reference image.
    This produces a reference image that depicts the target visual element without extraneous background or context, completing the quadruplet (source video, target video, reference image, instruction).

    \item \textbf{Quality Filtering.}
    A VLM-based filter (Gemini) evaluates each quadruplet on reference image clarity, reference-edit consistency, and instruction accuracy.
    Samples falling below quality thresholds are removed.
    This yields ${\sim}$700K single-reference editing quadruplets spanning diverse edit types including addition, replacement, background change, removal, and style transfer.
\end{enumerate}

\noindent\textbf{Multi-Reference Extension.}
Multi-reference editing data is constructed by performing two rounds of instruction-based editing.
Starting from the single-reference editing quadruplets produced above, we apply an additional instruction-based edit to obtain a second editing layer, which enables extracting a second reference image:
\begin{enumerate}[leftmargin=*,nosep]
    \item \textbf{Second-Round Instruction Editing.}
    Given a single-reference quadruplet (source video $V_s$, target video $V_t$, reference image $I_1$, instruction $c_1$), we apply a new instruction-based edit to $V_t$, producing a further-edited video $V_{t'}$ with a new editing instruction $c_2$.

    \item \textbf{Second Reference Extraction.}
    The newly edited video $V_{t'}$ and the previous target video $V_t$ now form a new (source, target) pair.
    We apply the same single-reference extraction pipeline described above (first-frame extraction, Gemini-based reference prompt generation, and Qwen Image Edit-based reference extraction) to obtain a second reference image $I_2$.

    \item \textbf{Multi-Reference Assembly.}
    This produces a multi-reference editing sample: (source video $V_s$, final target video $V_{t'}$, reference images $\{I_1, I_2\}$, composite instruction).

    \item \textbf{Quality Filtering.}
    A VLM-based quality filter removes samples with artifacts, inconsistent references, or conflicting edits.
    This produces ${\sim}$30K multi-reference video editing samples.
\end{enumerate}

\section{Subject-to-Video Data Construction}
\label{app:pipeline_ip}

We construct subject-to-video (S2V) training data from large-scale internal video collections through an automated subject detection, clustering, and pairing pipeline.
The goal is to produce training tuples of (reference subject image, target video, caption) where the reference subject appears in a different video than the target, preventing trivial copy-paste solutions.

The pipeline proceeds through six stages:
\begin{enumerate}[leftmargin=*,nosep]
    \item \textbf{Frame Sampling.}
    We uniformly sample 10+ keyframes from each video clip in the source collection (${\sim}$1M clips).

    \item \textbf{Per-Frame Subject Detection.}
    Qwen3-VL-8B~\cite{qwen3vl} performs open-vocabulary subject detection on each sampled frame, producing bounding boxes and confidence scores for all detected subjects (persons, animals, objects).
    Low-confidence detections ($<$0.09) are discarded.

    \item \textbf{Subject Cropping \& Embedding.}
    Each detected subject is cropped from its frame and encoded into a 768-dimensional identity embedding using FG-CLIP~\cite{xie2025fg}, a fine-grained visual encoder optimized for subject identity representation.

    \item \textbf{Cross-Video Identity Clustering.}
    Subject embeddings are clustered using K-means with cosine similarity to group instances of the same identity across different video clips.
    An additional Qwen-VL-based identity verification step filters false positives from the clusters.

    \item \textbf{Quality Filtering.}
    We apply multi-dimensional filtering to remove low-quality subject images: blur detection (Laplacian variance and learned blur scores), text overlay detection, occlusion detection, face/body consistency checks, and aesthetic scoring.
    Only subjects passing all quality gates are retained.

    \item \textbf{Caption Generation \& Formatting.}
    For each valid subject-video pair, a VLM (Gemini / Qwen-VL) generates a subject-driven caption in the format ``Based on the [subject] in Image-1, generate a video where\ldots,'' explicitly referencing the subject image.
    The final output is formatted as (reference image, target video, subject-conditioned caption) training tuples.
\end{enumerate}

\noindent
A key design choice is \emph{cross-video pairing}: the reference image is always drawn from a different video clip than the target.
This ensures the model cannot learn to trivially copy the reference into the target and must instead learn genuine identity-preserving generation.
The pipeline produces ${\sim}$230K single-reference S2V pairs.
An additional ${\sim}$100K high-quality multi-reference S2V samples are constructed by pairing multiple subject references with the same target video.

\section{Additional Qualitative Results}
\label{app:additional_qualitative}

We present additional qualitative results showcasing TIDE's capabilities across diverse editing types.

\subsection{Extended Comparisons on TIDE-Bench}
\label{app:tidebench_extended}

\begin{figure}[t]
\centering
\includegraphics[width=\textwidth]{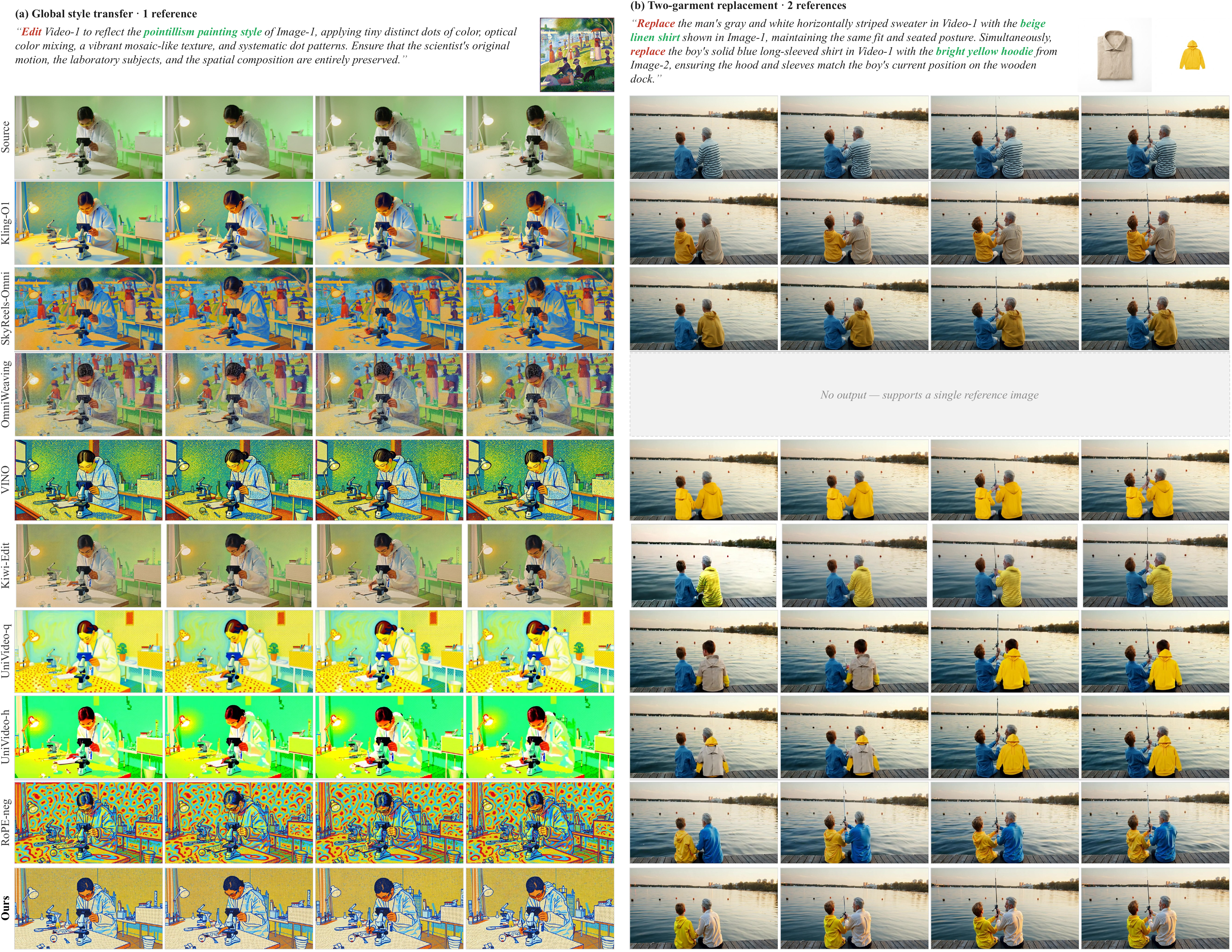}
\caption{\textbf{Additional qualitative comparisons on TIDE-Bench.} (a) Single-reference global style transfer to a pointillism appearance. (b) Two-reference garment replacement, where each reference must be bound to the correct subject. The released OmniWeaving TIV2V inference interface used in our evaluation accepts one reference image, so no result is reported for the two-reference case in (b).}
\label{fig:tidebench_cases2}
\end{figure}

Beyond the two cases shown against the full method pool in the main paper (Figure~\ref{fig:tidebench_qualitative}), Figure~\ref{fig:tidebench_cases2} adds two further cases at the two extremes of the benchmark.
Rows run from the source video through the closed-source and open-source systems to the RoPE-Neg variant and TIDE, four frames are sampled uniformly per clip, and the reference images sit beside the instruction with the phrases they depict marked.

\noindent\textbf{Qualitative analysis.}
As illustrated in Figure~\ref{fig:tidebench_cases2}, TIDE produces the most faithful result in both the global style-transfer and multi-reference garment-replacement cases.
For style transfer, it applies the referenced pointillism appearance while better preserving the scientist, laboratory content, motion, and overall spatial composition.
For garment replacement, it accurately assigns the two reference garments to different people and maintains both identities throughout the video, whereas the compared methods exhibit incomplete replacement, reference confusion, or visible appearance drift.
These cases demonstrate that explicit per-token task identifiers improve both reference fidelity and reference-to-target binding without degrading source-scene preservation.

\subsection{Multi-Reference Video Editing Cases}
\label{app:multiref_cases}

\begin{figure}[t]
\centering
\includegraphics[width=\textwidth]{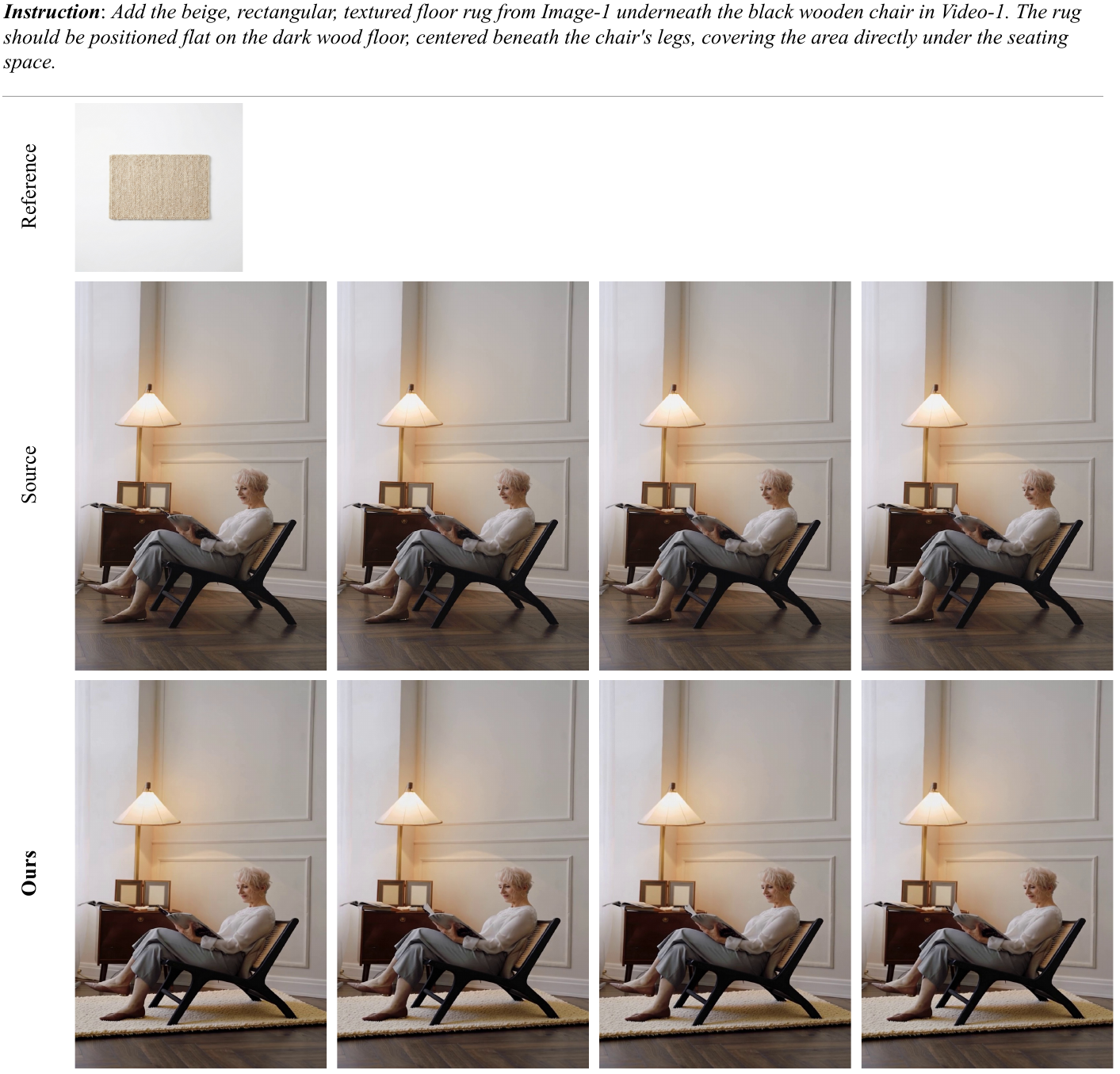}
\caption{Single-reference object addition on TIDE-Bench.}
\label{fig:case_add}
\end{figure}

\begin{figure}[t]
\centering
\includegraphics[width=\textwidth]{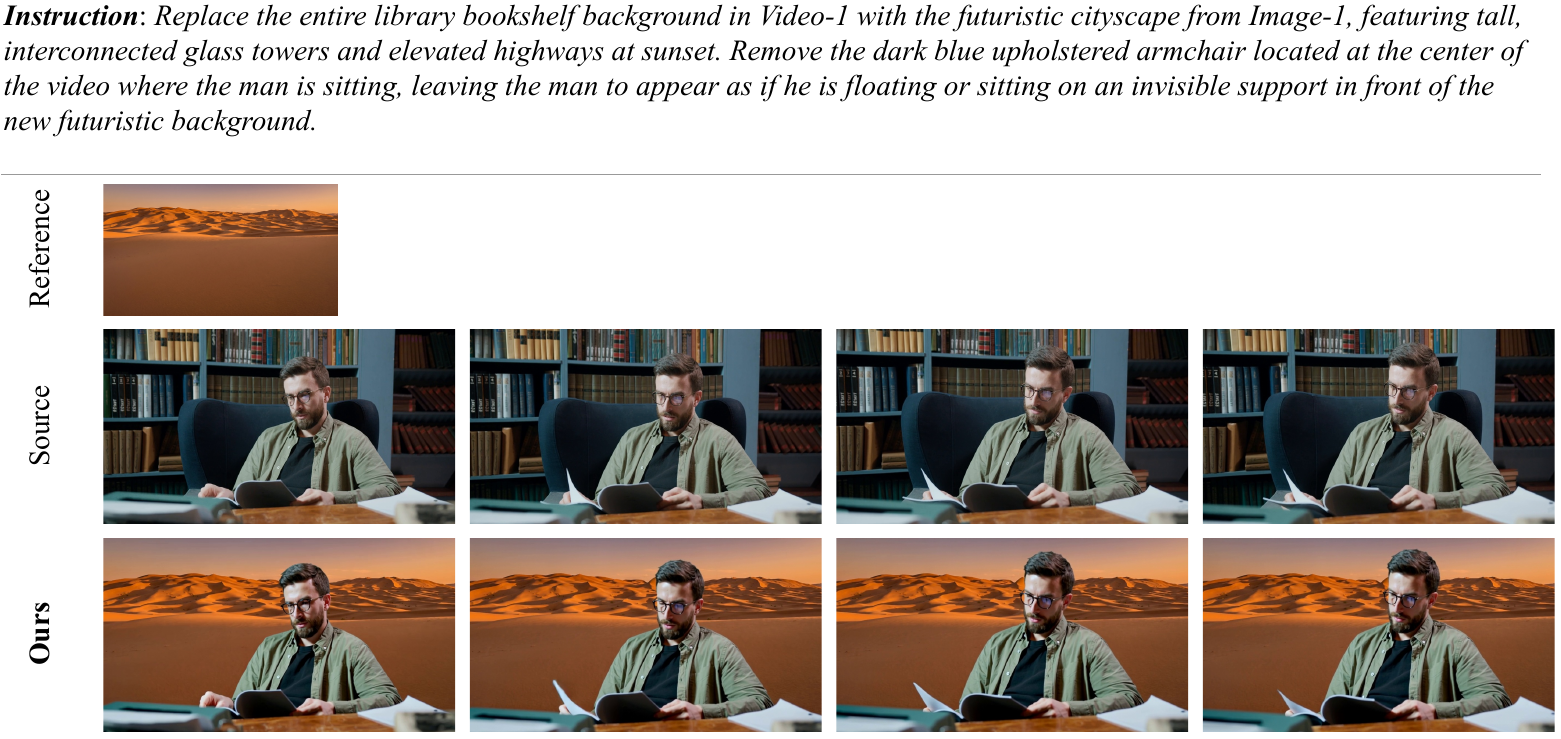}
\caption{Multi-reference editing: background change + object removal.}
\label{fig:case_bg_remove}
\end{figure}

\begin{figure}[t]
\centering
\includegraphics[width=\textwidth]{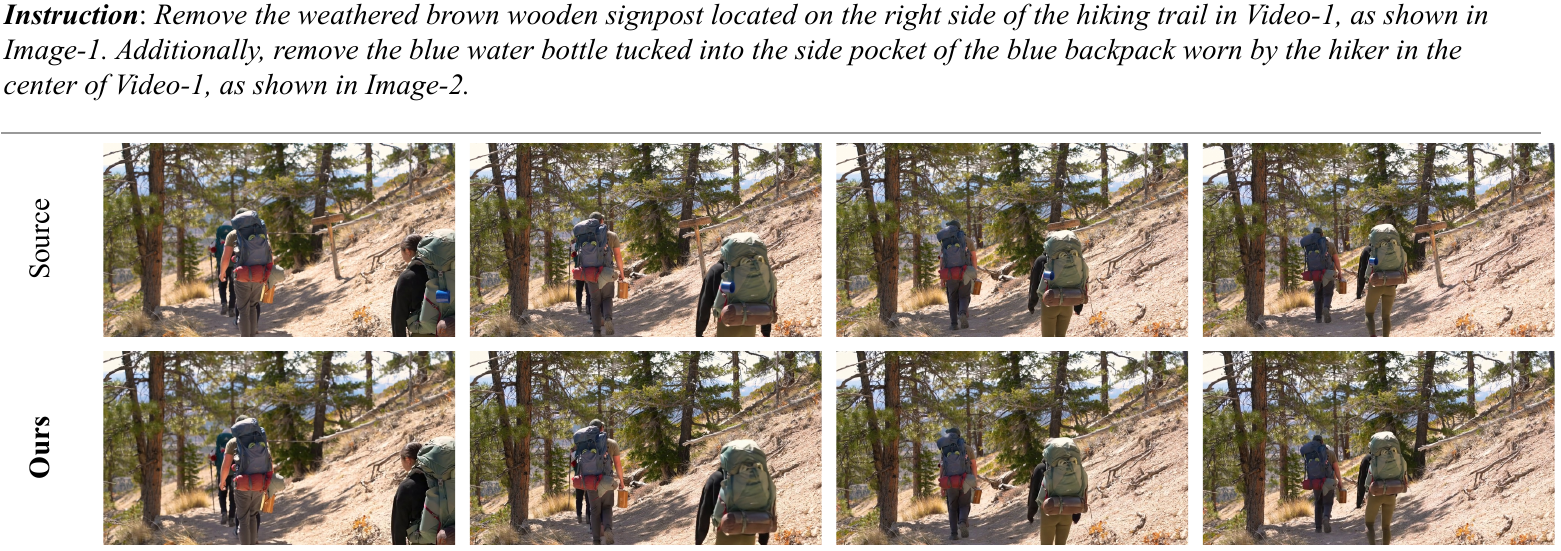}
\caption{Multi-reference editing: dual object removal.}
\label{fig:case_remove_remove}
\end{figure}

\begin{figure}[t]
\centering
\includegraphics[width=\textwidth]{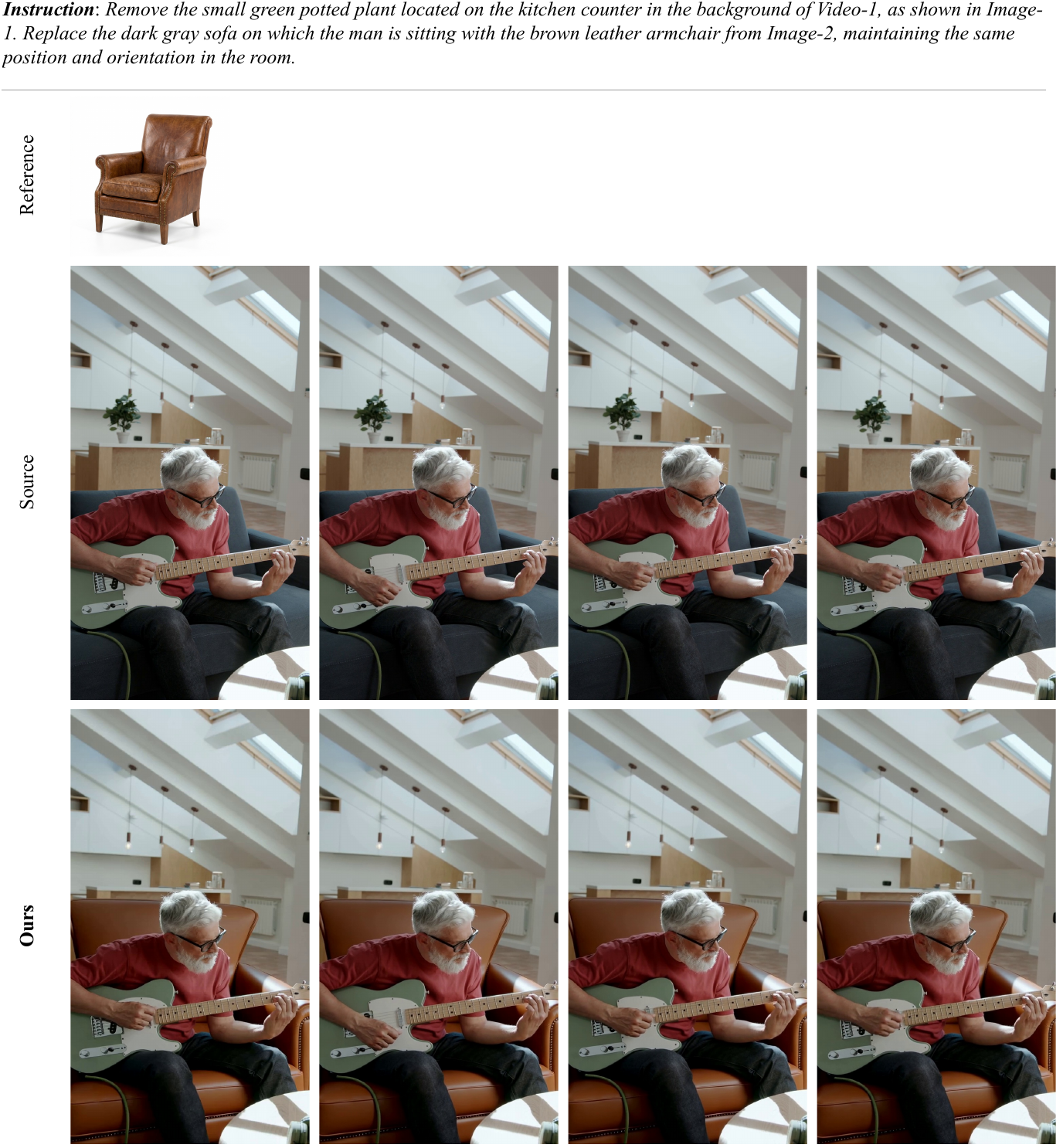}
\caption{Multi-reference editing: object removal + replacement.}
\label{fig:case_remove_replace}
\end{figure}

Figures~\ref{fig:case_add} to~\ref{fig:case_remove_replace} illustrate TIDE's multi-reference video editing across different operation combinations, including single-reference addition, background change with removal, dual object removal, and object removal with replacement.
Across these cases, TIDE accurately executes every requested operation, restores the supplied references with high fidelity, and preserves unedited subjects and scene regions, demonstrating robust control over both individual and compound edits.

\subsection{Subject-to-Video Generation Cases}
\label{app:s2v_cases}

\begin{figure}[t]
\centering
\includegraphics[width=0.70\textwidth]{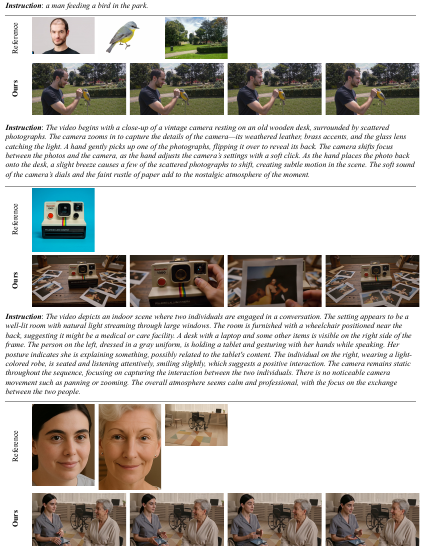}
\caption{Subject-to-video generation on OpenS2V with human and object references.}
\label{fig:case_s2v}
\end{figure}

As shown in Figure~\ref{fig:case_s2v}, TIDE faithfully preserves the visual identities of both human and object references while accurately following the text prompt.
The generated results maintain distinct subject appearances and coherent interactions over time, further demonstrating strong reference consistency and overall video quality.

\end{document}